\documentclass[11pt,a4paper]{article} 

\makeatletter
\providecommand*{\input@path}{}
\edef\input@path{{./}{2020_emnlp_storyteller/}\input@path}
\makeatother

\usepackage[hyperref]{style/emnlp2020}
\usepackage{times}
\usepackage{latexsym}
\usepackage{float}
\usepackage{enumitem}
\usepackage{amsmath}
\usepackage{amssymb}
\usepackage{textcomp}
\usepackage{booktabs}
\usepackage{array}
\usepackage{bbm}
\usepackage{url}
\usepackage{svg}
\usepackage{xcolor}
\usepackage{soul}
\usepackage{tcolorbox}
\usepackage{seqsplit}
\usepackage{multirow}
\usepackage{adjustbox}
\usepackage{makecell}
\usepackage{csquotes}
\usepackage{etoolbox}
\usepackage{tikz}
\usepackage{pifont}
\makeatletter
\patchcmd{\@maketitle}{\@author}{\@author\show\@thanks}{}{}
\makeatother

\usetikzlibrary{calc}
\usetikzlibrary{arrows}
\usetikzlibrary{arrows.meta}
\usetikzlibrary{decorations}
\usetikzlibrary{decorations.text}
\usetikzlibrary{decorations.pathmorphing}
\usetikzlibrary{decorations.pathreplacing}

\def\aclpaperid{1928}

\newif\ifcomment
\aclfinalcopy
\ifaclfinal
\commenttrue  
\newcommand{\storium}[0]{\textsc{storium}}
\else
\commenttrue  
\newcommand{\storium}[0]{\textsc{hush}}
\fi

\ifcomment
\newcommand{\npcomment}[1]{\textcolor{brown}{\bf \small [ #1 --Nanyun]}}
\else
\newcommand{\npcomment}[1]{}
\fi

\usepackage{mdwlist}
\usepackage{latexsym}
\usepackage{nicefrac}
\usepackage{booktabs}
\usepackage{amsfonts}
\usepackage{bold-extra}
\usepackage{amsmath}
\usepackage{dsfont}
\usepackage{amssymb}
\usepackage{bm}
\usepackage{graphicx}
\usepackage{mathtools}
\usepackage{microtype}
\usepackage{multirow}
\usepackage{multicol}
\usepackage{url}
\usepackage{latexsym,comment}

\ifcomment
\newcommand{\micomment}[1]{\textcolor{red}{\bf \small [#1 --Mohit]}}
\newcommand{\kkcomment}[1]{\textcolor{blue}{\bf \small [#1 --Kalpesh]}}
\newcommand{\tvcomment}[1]{\textcolor{green}{\bf \small [#1 --Tu]}}
\newcommand{\nacomment}[1]{\textcolor{orange}{\bf \small [#1 --Nader]}}
\newcommand{\swcomment}[1]{\textcolor{purple}{\bf \small [#1 --Shufan]}}
\else
\newcommand{\micomment}[1]{}
\newcommand{\tvcomment}[1]{}
\newcommand{\kkcomment}[1]{}
\newcommand{\nacomment}[1]{}
\newcommand{\swcomment}[1]{}
\fi

\newcommand{\gem}[1]{\mbox{\textsc{gem}}}

\newcommand{\hidetext}[1]{}
\newcommand{\ignore}[1]{}

\newcommand{\smallurl}[1]{ \begin{tiny}\url{#1}\end{tiny}}

\definecolor{lightblue}{HTML}{3cc7ea}
\definecolor{grey}{rgb}{0.95,0.95,0.95}
\definecolor{ceil}{rgb}{0.57, 0.63, 0.81}

\newcommand{\bleu}[0]{\textsc{bleu}}
\newcommand{\rouge}[0]{\textsc{rouge}}
\newcommand{\rougel}[0]{\textsc{rouge-l}}
\newcommand{\rougew}[0]{\textsc{rouge-w}}
\newcommand{\storymetric}[0]{\textsc{user}}
\newcommand{\storymetricfull}[0]{User Story Edit Ratings}
\newcommand{\writingprompts}[0]{\texttt{r/writingprompts}}
\newcommand{\roleplayerguild}[0]{{\color{blue!50!black}\texttt{roleplayerguild}}}
\newcommand{\bmat}[1]{\text{\textbf{#1}}}
\newcommand{\bvec}[1]{\boldsymbol{#1}}
\newcommand{\namedref}[2]{\hyperref[#2]{#1~\ref*{#2}}}
\newcommand{\sectionref}[1]{\namedref{Section}{sec:#1}}
\newcommand{\tableref}[1]{\namedref{Table}{table:#1}}
\newcommand{\figureref}[1]{\namedref{Figure}{fig:#1}}
\newcommand{\appendixref}[1]{\namedref{Appendix}{appendix:#1}}
\newcommand{\footnoteref}[1]{\textsuperscript{\ref{footnote:#1}}}
\newcommand{\cmark}{{\color{green!70!black}\ding{51}}}
\newcommand{\xmark}{{\color{red}\ding{55}}}

\definecolor{introcolor}{RGB}{0,0,128}
\definecolor{challengecolor}{RGB}{128,0,0}
\definecolor{strengthcolor}{RGB}{128,0,128}
\definecolor{charactercolor}{RGB}{128,128,128}
\definecolor{entrycolor}{RGB}{64,128,64}
\definecolor{preventrycolor}{RGB}{128,128,0.0}
\definecolor{titlecolor}{RGB}{64,13,13}
\definecolor{descriptioncolor}{RGB}{0.0,128,128}
\definecolor{lightgray}{gray}{0.94}

\newtcbox{\cbox}[2][red]{
  tcbox raise=-3pt,
  nobeforeafter,
  arc=0pt,
  outer arc=0pt,
  colback=#1!40!grey,
  colframe=#1!50!gray,
  colupper=white,
  boxsep=0pt,
  left=0pt,
  right=0pt,
  top=#2,
  bottom=#2,
  boxrule=0pt,
  bottomrule=1pt,
  toprule=1pt,
  valign=center
}

\makeatletter

\newcommand{\defhighlighter}[2][]{%
  \tikzset{every highlighter/.style={color=#2!40!, #1}}%
}

\defhighlighter{yellow}
\newif\ifhighlight
\highlighttrue

\newcommand{\highlight@DoHighlight}{
  \ifhighlight
    \fill[
      decoration={bent, amplitude=0.5pt},
      outer sep=-1em,
      inner sep=0pt,
      every highlighter,
      this highlighter,
      decorate
      ]
      ($(begin highlight)+(0,6pt)$) rectangle ($(end highlight)+(0,-2pt)$);
  \fi
}

\newcommand{\highlight@BeginHighlight}{
  \coordinate (begin highlight) at (0,0);
}

\newcommand{\highlight@EndHighlight}{
  \coordinate (end highlight) at (0,0);
}

\newdimen\highlight@previousx
\newdimen\highlight@previousy
\newdimen\highlight@currentx
\newdimen\highlight@currenty

\def\SOUL@fancyhlpreamble{%
  \begin{tikzpicture}[overlay, remember picture]
    \highlight@BeginHighlight
    \highlight@EndHighlight
  \end{tikzpicture}%
}%

\def\SOUL@fancyhlpostamble{%
  \begin{tikzpicture}[overlay, remember picture]
    \highlight@EndHighlight
    \highlight@DoHighlight
  \end{tikzpicture}%
}%

\def\SOUL@fancyhleveryhyphen{%
  \discretionary{%
    \SOUL@setkern\SOUL@hyphkern
    \SOUL@sethyphenchar
    \tikz[overlay, remember picture] \highlight@EndHighlight;%
  }{%
  }{%
    \SOUL@setkern\SOUL@charkern
  }%
}%

\def\SOUL@fancyhleveryexhyphen#1{%
  \SOUL@setkern\SOUL@hyphkern
  \hbox{#1}%
  \discretionary{%
    \tikz[overlay, remember picture] \highlight@EndHighlight;%
  }{%
  }{%
    \SOUL@setkern\SOUL@charkern
  }%
}%

\def\SOUL@fancyhleverysyllable{%
  \begin{tikzpicture}[overlay, remember picture]
    \path let \p0 = (begin highlight), \p1 = (end highlight), \p2 = (0, 0) in \pgfextra
      \global\highlight@previousx=\x0
      \global\highlight@previousy=\y0
      \global\highlight@currentx =\x1
      \global\highlight@currenty =\y2
    \endpgfextra (0,0) ;
    \ifdim\highlight@currenty < \highlight@previousy
      \ifdim\highlight@currentx > \highlight@previousx
        \highlight@DoHighlight
      \else
        \highlight@EndHighlight
      \fi
      \highlight@BeginHighlight
    \fi
  \end{tikzpicture}%
  \the\SOUL@syllable
  \SOUL@setkern\SOUL@charkern
  \tikz[overlay, remember picture] \highlight@EndHighlight;%
}%

\DeclareRobustCommand*\highlightdriver[1][]{%
  \tikzset{this highlighter/.style={#1!40!}}%
  \SOUL@setup
  \let\SOUL@preamble\SOUL@fancyhlpreamble
  \let\SOUL@postamble\SOUL@fancyhlpostamble
  \let\SOUL@everyhyphen\SOUL@fancyhleveryhyphen
  \let\SOUL@everyexhyphen\SOUL@fancyhleveryexhyphen
  \let\SOUL@everysyllable\SOUL@fancyhleverysyllable
  \SOUL@
}

\DeclareRobustCommand*\highlight[2][yellow]{%
  \highlightdriver[#1]{#2}%
}

\DeclareRobustCommand*\highlightinline[2][yellow]{%
  \highlighttrue%
  \highlightdriver[#1]{#2}%
  \highlightfalse%
  \makebox[0pt][r]{\highlightdriver[#1]{#2}}%
}

\makeatother

\title{\storium: A Dataset and Evaluation Platform for Machine-in-the-Loop Story Generation}

\author{Nader Akoury\textsuperscript{\textnormal{\ensuremath{\dagger\ast}}}
Shufan Wang\textsuperscript{\textnormal{\ensuremath{\dagger}}}
Josh Whiting\textsuperscript{\textnormal{\ensuremath{\ddagger}}}
Stephen Hood\textsuperscript{\textnormal{\ensuremath{\ddagger}}} \\
\textbf{Nanyun Peng\textsuperscript{\textnormal{\S}}
Mohit Iyyer\textsuperscript{\textnormal{\ensuremath{\dagger}}}} \\
\textsuperscript{\ensuremath{\dagger}}University of Massachusetts Amherst
\textsuperscript{\ensuremath{\ddagger}}Storium
\textsuperscript{\S}University of California Los Angeles \\
\texttt{\normalsize\{nsa,shufanwang,miyyer\}@cs.umass.edu} \\
\texttt{\normalsize\{josh,stephen\}@storium.com} \\
\texttt{\normalsize violetpeng@cs.ucla.edu}
}

\setitemize{itemsep=2pt,topsep=2pt,parsep=0pt,partopsep=5pt}
\setenumerate{itemsep=2pt,topsep=2pt,parsep=0pt,partopsep=5pt,itemindent=13pt}

\begin{document}
\maketitle

\begin{abstract}
  Systems for \emph{story generation} are asked to produce plausible and
  enjoyable stories given an input context. This task is underspecified, as a
  vast number of diverse stories can originate from a single input. The large
  output space makes it difficult to build and evaluate story generation
  models, as (1) existing datasets lack rich enough contexts to meaningfully
  guide models, and (2) existing evaluations (both crowdsourced and automatic)
  are unreliable for assessing long-form creative text. To address these
  issues, we introduce a dataset and evaluation platform built from \storium,
  an online collaborative storytelling community. Our author-generated dataset
  contains 6K lengthy stories (125M tokens) with fine-grained natural
  language annotations (e.g., character goals and attributes) interspersed
  throughout each narrative, forming a robust source for guiding models. We
  evaluate language models fine-tuned on our dataset by integrating them onto
  \storium, where \emph{real} authors can query a model for suggested story
  continuations and then edit them. Automatic metrics computed over these
  edits correlate well with both user ratings of generated stories and
  qualitative feedback from semi-structured user interviews. We release both
  the \storium\ dataset and evaluation platform to spur more principled
  research into story generation. 
\end{abstract}

\section{Introduction}
\label{sec:intro}

\begin{figure*}
  \centering
  \resizebox{\textwidth}{!}{%
    \input{figs/scene_outline.tex}
  }
  \caption{
    A high-level outline of our dataset and platform. In this example from a
    real \storium\ game, the \cbox[charactercolor]{1.925pt}{character}
    \cbox[charactercolor]{2.75pt}{\textsc{adira makarova}} uses the
    \cbox[strengthcolor]{0.925pt}{strength card}
    \cbox[strengthcolor]{2.75pt}{\textsc{deadly aim}}  to
    \cbox[challengecolor]{2.75pt}{\textsc{disrupt the germans}}, a
    \cbox[challengecolor]{0.925pt}{challenge card}. Our model conditions on the
    natural language annotations in the \cbox[introcolor]{1.925pt}{scene
    intro}, \cbox[challengecolor]{0.925pt}{challenge card},
    \cbox[strengthcolor]{0.925pt}{strength card}, and
    \cbox[charactercolor]{1.925pt}{character}, along with the text of the
    \cbox[preventrycolor]{0.925pt}{previous scene entry} (not shown) to
    generate a suggested story continuation. Players may then edit the model
    output, by \highlightinline[green]{adding} or
    \highlightinline[red]{deleting} text, before publishing the entry. We
    collect these edits, using the \highlightinline[yellow]{matched} text as
    the basis of our \storymetric\ metric. New models can be added to the
    platform by simply implementing four methods: \texttt{startup},
    \texttt{shutdown}, \texttt{preprocess}, and \texttt{generate}.
  }
  \label{fig:example_cards}
\end{figure*}

Fiction writers express their creativity through both low-level linguistic
choices and discourse-level sequencing of narrative elements (e.g., plot events
and character development). Unlike more constrained text generation tasks, such
as translation or summarization, fiction writing allows for almost infinite
creative freedom, which budding authors often find cognitively
overwhelming~\citep{rose1980rigid}. Machine-in-the-loop
storytelling~\citep{Clark2018CreativeWW}, in which an author obtains
automatically generated sentences or paragraphs when stuck with writer's block,
lowers the barrier to entry for creative writing~\citep{roemmele2015creative}.
To spur research in this area, we partner with
\storium,\footnote{\ifaclfinal\url{https://storium.com}\else To preserve author
anonymity, we use \storium\ in place of the actual platform name in this
submission.\fi} an online collaborative storytelling platform, to introduce a
new dataset and evaluation methodology for story generation.

The open-endedness of story writing does not just pose a barrier to humans---it
also presents a challenge for building and evaluating computational models.
Prior work relies on datasets that are either too artificial to generalize to
long-form stories, such as the crowdsourced
ROCStories~\citep{mostafazadeh2016corpus} corpus, or too unconstrained, as in
the \writingprompts\ dataset~\citep{fan2018hierarchical}, which pairs
medium-length stories with short prompts. Furthermore, lack of standardized
evaluation makes measuring progress difficult: most prior work evaluates
outputs using a combination of simple automatic metrics not designed for
long-form creative text generation (e.g., \bleu\ and \rouge\ against a single
reference) and crowdsourced ratings
\cite{McIntyre2009LearningTT,yao2019plan,fan2019strategies} that preclude
evaluating long-form narratives.

We address these limitations by (1) collecting a dataset of stories
(\sectionref{dataset}) containing fine-grained structural annotations written
in natural language, and (2)
providing a platform for evaluating models in a machine-in-the-loop setting by
allowing real \storium\ authors to interact with the generated stories
(\sectionref{methodology}). Our dataset contains nearly 6K longform stories
(125M tokens) written by \storium\ authors, each of which is broken into
discourse-level scene entries annotated with narrative elements, such as
character goals or abilities. Conditioning story generation models on this
information thus imposes loose constraints on what the model should produce,
compared to unstructured datasets such as \writingprompts, and also enables
modeling of narrative planning processes.

We fine-tune large-scale pretrained language models on our dataset
(\sectionref{model}) and integrate them with the \storium\ platform, where
authors can query a model for the next few sentences in their story and then
edit the resulting text to their liking. We devise a metric (inspired by
\rouge) on top of these edits that measures how much of the generated text is
preserved in the post-edited version, and discover that this metric correlates
with Likert judgments of linguistic properties such as relevance and coherence.
Detailed analyses of the edits (\sectionref{analysis}), including
semi-structured interviews with \storium\ users, suggests that generating text
\emph{relevant} to the current story context is the most important open problem
in this area. We publicly release both the \storium\ dataset and user-facing
evaluation platform to facilitate future research on story
generation.\ifaclfinal\footnote{\url{https://storium.cs.umass.edu}}\else\footnote{Data
and evaluation platform to be released after review.}\fi

\section{\storium\ Dataset}
\label{sec:dataset}

Our \storium\ dataset derives from an online collaborative storytelling
community that provides rich metadata useful for guiding computational
storytelling systems. In this section, we describe how the structural elements
of \storium\ stories fit together, and verify via an annotation task that this
metadata indeed influences the text of the stories. Finally, we use neural
topic models to highlight the thematic content and narrative sequencing of
\storium.

\begin{table*}[ht]
  \centering
  \scalebox{0.975}{\begin{tabular}{lrrccc}
\toprule
  \bfseries Dataset & \bfseries \# Stories & \bfseries \# Tokens per Story & \bfseries Prompts & \bfseries Turns & \bfseries Annotations \\
  \midrule
  \roleplayerguild & 1,439 & 3,079\parbox{0pt}{$^\ast$} & \xmark & \cmark & \xmark \\
  PG-19 & 28,752 & 68,973 & \xmark & \xmark & \xmark \\
  ROCStories & 98,156 & 88 & \cmark & \xmark & \xmark \\
  \writingprompts & 303,358 & 735 & \cmark & \xmark & \xmark \\
  \midrule
  \storium & 5,743 & 19,278 & \cmark\parbox{0pt}{$^\dagger$} & \cmark & \cmark \\
  \bottomrule
\end{tabular}
}
  \caption{While \storium\ has fewer stories than other popular story datasets,
  each story is considerably longer and contains natural language annotations
  to guide story generation. $^\ast$We combine character and action sets to
  determine average story length. $^\dagger$We count narrator actions
  introducing challenges and locations as prompts.}
  \label{table:dataset-comparison}
\end{table*}

\subsection{\storium: Gamified Storytelling}
The \storium\ platform enables a small group of users to collaboratively write
a single story by transforming the writing process into a turn-based game. In
each game, one player acts as the \textit{narrator}, while other players take
on the role of individual \textit{characters} within the story (e.g.,
\cbox[charactercolor]{2.75pt}{\textsc{adira makarova}} in
\figureref{example_cards}). Stories unfold through a series of high-level
\textit{scenes} that consist of multiple short \textit{entries}, each of which
is written from the perspective of a character (or the narrator). Scenes
commonly revolve around \textit{challenges} (e.g.,
\cbox[challengecolor]{2.75pt}{\textsc{disrupt the germans}}), that the
characters tackle within the text of their entries; to help address these
challenges, each character has access to a set of \textit{cards} (e.g.,
\cbox[strengthcolor]{2.75pt}{\textsc{deadly aim}}, a
\cbox[strengthcolor]{0.925pt}{strength card}) that define various properties
such as strengths, weaknesses, items, and goals. The narrator moves the story
forward by introducing new challenges, locations, and characters, in the form
of cards. These are either created from scratch by the narrator or selected
from a predefined \textit{world} that contains a common set of story elements.
Collectively, the cards played form a set of structural natural language
annotations that guide the story being written.

\paragraph{Dataset details:}
We collect 5,743 publicly available stories written on \storium\ from January
2015 to August 2019. We reserve 569 stories for validation and 570 stories for
test --- carefully ensuring an 8:1:1 split with respect to both the number of
stories and tokens. Altogether, the stories are broken down into 25,092 scenes
with 448,264 individual scene entries (126,041,738 tokens), conditioned on
232,596 cards, 204,698 of which are unique.
\begin{table}[H]
  \centering
  \normalsize
  \begin{tabular}{lr}
\toprule
Stories & 5,743 \\
Authors & 30,119 \\
Characters & 25,955 \\
Scenes & 25,092 \\
Scene Entries & 448,264 \\
Cards Played & 232,596 \\
Average Tokens$^\ast$ (per Entry) & 247 \\
Average Tokens$^\ast$ (per Story) & 19,278 \\
Total Tokens$^\ast$ (Entries \& Cards) & 126,041,738 \\
\bottomrule
\end{tabular}

  \caption{An overview of our dataset, which contains long stories, broken down
  into scene entries, with structural annotations in the form of cards played
  to guide the narrative. $^\ast$We count tokens as contiguous spans of either
  alphanumeric or non-alphanumeric symbols.}
  \label{table:dataset-condensed}
\end{table}

\paragraph{Cards influence entry text: }
\storium\ does not force players to relate their written entries to selected
cards or challenges, instead relying on game conventions to guide user
behavior. To validate whether the structural metadata influences story text, we
conduct a small-scale annotation of 235 scene entries, where we ask
annotators\footnote{The annotators were NLP graduate students.} to provide
binary judgments for (1) whether the card played influences the scene entry,
and (2) if the scene entry addresses the current challenge. We find that 77\%
of scene entries reference the played cards, and 80\% address the current
challenge (\tableref{card-influence}).

\paragraph{Related datasets:}
Prior story generation papers have frequently focused on the
ROCStories~\citep{mostafazadeh2016corpus} and
\writingprompts~\citep{fan2018hierarchical} datasets. While \storium\ has
comparatively fewer stories than these datasets, our stories are over an order
of magnitude longer (Table~\ref{table:dataset-comparison}). Rather than
containing a single short prompt to start the story, our stories on average
contain 14 narrator prompts per story, with 41 natural language annotations
which describe character goals, attributes, and key items useful for
conditioning story generation models.\footnote{While \citet{fan2019strategies}
extract internal structure via SRL, this is not inherent to the dataset, and
can be applied to other datasets, including our own.} Like \storium, the
stories in \roleplayerguild~\citep{louis_deep_2018} are also formed from
collaborative storytelling turns via a  role-playing game, though this dataset
lacks any prompts or annotations. Finally, datasets consisting of novels and
other fiction, like PG-19~\citep{Rae2020Compressive}, provide long-form
narratives without explicit structure to constrain generation.

\subsection{Common Themes and Story Arcs}
To provide insight into common narrative themes and substructures within our
dataset, we train a neural topic model on text from entries and challenges and
analyze the resulting topics and their transitions. 

\subsubsection{Topic model specification} 
Our topic model is a simplified version of the \emph{relationship modeling
network} (\textsc{rmn}) proposed by~\citet{Iyyer16}.\footnote{Preliminary
experiments with LDA~\citep{blei2003latent} yielded less coherent topics, which
is consistent with evaluations in~\citet{Iyyer16}.} As in the \textsc{rmn}, our
model relies on dictionary learning to compute topics; however, it models each
entry and challenge independently, instead of considering the temporal order of
scenes through recurrence. We ignore the temporal component because \storium\
contexts do not neatly fit into a chronologically-ordered timeline (e.g.,
entries within a single scene may not depend on each other). Building a
specialized topic model for this data is beyond the scope of this work.

Concretely, given an input text (either an entry or a challenge), we first
encode it by computing an average of pretrained GloVe\footnote{glove.840B.300d}
embeddings $\bvec{x}$. Next, we compute the dot product between $\bvec{x}$ and
each row of a global \emph{dictionary matrix} $\bmat{R}$. Intuitively, each row
of $\bmat{R}$ is a vector representation of an individual topic. These row-wise
dot products are converted to a probability distribution via a softmax function
and then used to compute a weighted average $\bvec{r}$ of the dictionary rows,
which is then trained through a contrastive max-margin loss to reconstruct the
input vector $\bvec{z}$. At test time, the dictionary rows are interpreted by
their nearest neighbors (using cosine distance) in the GLoVe word embedding
space.\footnote{We encourage interested readers to see~\citet{Iyyer16} for more
details. The only difference between our setup and theirs
is that we directly use $\bvec{x}$ to compute the row weights without any
feed-forward or recurrent layers in between.}

\begin{table}[thb]
  \centering
  \small
  \begingroup
\setlength{\tabcolsep}{1ex}
\begin{tabular} {p{2cm}p{5cm}}
\toprule
\bfseries Worlds & \bfseries Topic words \\
\midrule
Fantasy Classic & rotunda, courtyard, staircase, foyer \\[2pt]
Urban Fantasy & analyze, investigate, analyse, uncover \\[2pt] 
The Mysterious Island & convoy, hiking, river, reconnaissance \\[2pt] 
Cyberpunk & synchronization, decryption, device, apparatus \\[2pt] 
Steampunk & freighter, crewmembers, cockpit, airship \\[2pt] 
The Heroes \newline Return & thine, fealty, uphold, valor \\[2pt] 
Medical Drama & tumor, ligament, laceration, mortem \\[2pt] 
Los Chicos \newline Malos & sublight, biosphere, aetheric, gravitational \\[2pt] 
The University & explanation, undergrad, spelling, reasoning \\[2pt] 
The 33 & melodramatic, reenactment, film, thriller \\[2pt] 
Scrapjack & brake, soldering, heater, corrosion \\  \bottomrule
\end{tabular}
\endgroup

  \caption{Topics with the highest relative importance for a sample of
  \storium\ worlds, which illustrate the diversity of the dataset.}
  \label{table:relative-impt}
\end{table}

\subsubsection{Examining topics and their transitions}
To explore the content of the \storium\ dataset, we train our model with 50
topics (i.e., $\bmat{R}$ has 50 rows) on the union of entry and challenge text.
\tableref{relative-impt} shows the most distinguishing topic (ranked by
relative importance) for a sample of different \storium\ worlds. These topics
illustrate the diversity of our dataset: topics range from science fiction
(\emph{Cyberpunk}, \emph{Steampunk}) to detective fiction (\emph{Urban
Fantasy}) and stories set in hospitals (\emph{Medical Drama}) and schools
(\emph{The University}). 

Following the methodology of~\citet{antoniak2019narrative}, we also examine
common local topic transitions between entries written by the \emph{same}
character across different scenes in a story. We compute the transition
probability from topic $A$ to topic $B$ by counting how many times $A$  and $B$
are the most probable topics for two consecutive entries, respectively, and
normalizing by the total number of occurrences of topic $A$.
\figureref{story_arcs} shows a topic transition diagram originating from a
weapons-related topic. In the \emph{Space Adventure} world, stories
progress into vehicle and technology-related topics, while in \emph{Fantasy
Classic}, they tend to transition to topics about valor instead. That said,
both of these worlds are not completely different, as they share a transition
topic associated with physical action.

\begin{figure}
  \centering
  \begin{tikzpicture}
  \tikzstyle{Topic} = [
    outer sep=0,
    inner sep=5pt,
    rounded corners,
    font={\sffamily\scriptsize\itshape},
    text=black!80!violet,
    text centered,
    text width=10ex,
  ]

  \tikzstyle{Bubble} = [
    decorate,
    fill=violet!10!,
    decoration={snake,amplitude=1pt,segment length=2mm}
  ]

  \tikzstyle{BubbleSize} = [
    x radius=1cm,
    y radius=0.5cm,
    yshift=-2pt
  ]

  \tikzstyle{Arrows} = [
    black,
    semithick,
    font=\tiny\sffamily\bfseries,
    double distance=5.25pt,
    postaction={
      decorate,
      decoration={
        text={#1},
        text align=center,
        text effects along path,
        text effects/every character/.style={text along path,yshift=-0.5pt,black!2!,inner sep=0.75pt}
      }
    }
  ]

  \tikzstyle{UrbanArrows} = [
    Arrows={#1},
    double=black!40!,
    -{Stealth[open,angle=45:1.5em,fill=teal!50!white]},
  ]
  \tikzstyle{ClassicArrows} = [
    Arrows={#1},
    double=black!40!,
    -{Stealth[open,angle=45:1.5em,fill=brown!50!white]},
  ]
  \tikzstyle{DoubleArrows} = [
    Arrows={#1},
    double=black!40!,
    -{
      Stealth[open,angle=45:1.5em,fill=teal!50!white,quick,sep=-12pt]
      Stealth[open,angle=45:1.5em,fill=brown!50!white,quick]
    },
  ]

  \tikzstyle{LungeArrows} = [
    black,
    semithick,
    font=\tiny\sffamily\bfseries,
    double distance=5.25pt,
    postaction={
      decorate,
      decoration={
        text={#1},
        text align=center,
        text effects along path,
        text effects/every character/.style={text along path,yshift=-2.5pt,black!2!,inner sep=0.75pt}
      }
    }
  ]

  \tikzstyle{LungeUrbanArrows} = [
    LungeArrows={#1},
    double=black!40!,
    -{Stealth[open,angle=45:1.5em,fill=teal!50!white]},
  ]
  \tikzstyle{LungeClassicArrows} = [
    LungeArrows={#1},
    double=black!40!,
    -{Stealth[open,angle=45:1.5em,fill=brown!50!white]},
  ]
  \tikzstyle{LungeDoubleArrows} = [
    LungeArrows={#1},
    double=black!40!,
    -{
      Stealth[open,angle=45:1.5em,fill=teal!50!white,quick,sep=-12pt]
      Stealth[open,angle=45:1.5em,fill=brown!50!white,quick]
    },
  ]

  \node[Topic] (topic1) {weapon combat melee};

  \node[Topic, anchor=north, yshift=-2em]
    (topic3) at (topic1.south)
    {lunge swerving uppercut};
  \node[Topic, anchor=east, xshift=-1em]
    (topic2) at (topic3.west)
    {fealty valor sword};
  \node[Topic, anchor=west, xshift=1em]
    (topic4) at (topic3.east)
    {reconnaissance convoy patrol};

  \node[Topic, anchor=north, yshift=-2em]
    (topic5) at (topic3.south)
    {mailing notify caller};

  \draw[ClassicArrows={weapon},/pgf/decoration/reverse path=true]
    ($(topic1.north west)!0.3!(topic1.south west)+(5pt,0)$)
    .. controls +(180:2em) and +(90:2em) ..
    ($(topic2.north)!0.55!(topic2.north)$);
  \draw[DoubleArrows={melee},/pgf/decoration/reverse path=true]
    ($(topic1.south west)!0.4!(topic1.north)$)
    .. controls +(180:2em) and +(135:2em) ..
    ($(topic3.west)!0.1!(topic3.south)$);
  \draw[UrbanArrows={combat}]
    ($(topic1.north east)!0.3!(topic1.south east)-(5pt,0)$)
    .. controls +(0:2em) and +(90:2em) ..
    ($(topic4.north)!0.0!(topic4.north)$);
  \draw[LungeDoubleArrows={lunge}]
    ($(topic3.north east)!0.5!(topic3.south)$)
    .. controls +(0:2em) and +(315:2em) ..
    ($(topic1.south east)!0.3!(topic1.north east)-(3.5pt,0)$);
  \draw[LungeUrbanArrows={reconnaissance},/pgf/decoration/reverse path=true]
    ($(topic4.north west)!0.75!(topic4.south east)$)
    .. controls +(315:3em) and +(0:3em) ..
    ($(topic5.north east)!0.75!(topic5.south east)-(5pt,0)$);
  \draw[LungeClassicArrows={lunge},/pgf/decoration/reverse path=true]
    ($(topic3.south)!0.1!(topic3.south)$)
    .. controls +(270:2em) and +(290:2em) ..
    ($(topic2.south)!0.3!(topic2.south)$);

  \draw [Bubble] (topic1) circle [BubbleSize];
  \node[Topic] (topic1) at (topic1) {weapon combat melee};

  \draw [Bubble] (topic2) circle [BubbleSize];
  \node[Topic] (topic2) at (topic2) {fealty valor sword};

  \draw [Bubble] (topic3) circle [BubbleSize];
  \node[Topic] (topic3) at (topic3) {lunge swerving uppercut};

  \draw [Bubble] (topic4) circle [BubbleSize];
  \node[Topic] (topic4) at (topic4) {reconnaissance convoy patrol};

  \draw [Bubble] (topic5) circle [BubbleSize];
  \node[Topic] (topic5) at (topic5) {mailing notify caller};

  \newtcbox{\legendbox}[3][red] {
    on line,
    nobeforeafter,
    arc=0pt,
    colupper=white,
    colback=#1!50!white,
    colframe=#1!50!black,
    before upper={
      \rule[-3pt]{0pt}{10pt}
    },
    boxrule=0.6pt,
    boxsep=0pt, 
    left=#2, 
    right=#3,
    top=0pt,
    bottom=0pt,
    bottomrule=0.6pt,
  }

  \node[Topic, font={\scriptsize\sffamily\bfseries}, text width=30ex]
    (legend) at ($(topic5.south)-(0,2em)$)
    {
      \legendbox[brown]{3.5pt}{3.5pt}{\textsc{fantasy classic}}
      \legendbox[teal]{2pt}{2pt}{\textsc{space adventure}}
    };

\end{tikzpicture}
  \caption{Example story arcs derived from the adjacency matrix of topic
  transitions over the text of entries (e.g., in \tcbox[
    on line,
    nobeforeafter,
    arc=0pt,
    top=0pt,
    bottom=0pt,
    left=2pt,
    right=2pt,
    boxrule=0.6pt,
    colupper=white,
    colback=brown!50!white,
    colframe=brown!50!black
  ]{\scriptsize\bfseries\textsc{\sffamily{fantasy classic}}}
  stories, the
  \tcbox[
    on line,
    nobeforeafter,
    arc=0pt,
    top=0pt,
    bottom=0pt,
    left=2pt,
    right=2pt,
    boxsep=0pt,
    boxrule=0pt,
    colback=violet!10!,
    colupper=black!80!violet,
    toprule=0pt,
    bottomrule=0pt,
  ]{\small\sffamily\textit{weapon, combat, melee}}
  topic is often followed by a
   transition, as denoted by
  \tcbox[
    on line,
    nobeforeafter,
    arc=0pt,
    top=1pt,
    bottom=0pt,
    left=2pt,
    right=2pt,
    boxsep=0pt,
    boxrule=0pt,
    colframe=black,
    colback=black!40!,
    colupper=black!2!,
    toprule=0.6pt,
    bottomrule=0.6pt,
  ]{\small\sffamily\bfseries{weapon}},
  to the
  \tcbox[
    on line,
    nobeforeafter,
    arc=0pt,
    top=0pt,
    bottom=0pt,
    left=2pt,
    right=2pt,
    boxsep=0pt,
    boxrule=0pt,
    colback=violet!10!,
    colupper=black!80!violet,
    toprule=0pt,
    bottomrule=0pt,
  ]{\small\sffamily\textit{fealty, valor, sword}}
  topic).}
  \label{fig:story_arcs}
\end{figure}

\section{Generating Scene Entries}
\label{sec:model}

\begin{figure*}[t]
  \centering
  \begin{tikzpicture}
  \definecolor{poscolor1}{RGB}{255,0,0}
  \definecolor{poscolor2}{RGB}{255,255,255}
  \definecolor{tokencolor}{RGB}{64,128,192}

  \tikzstyle{Basic} = [
    anchor=west,
    inner sep=2.5pt,
    outer sep=0,
    minimum height=\tokenwidth,
    minimum width=\tokenheight,
    text depth=1pt,
    text=black,
    font={\footnotesize\sffamily}
  ]

  \tikzstyle{Token} = [Basic, draw]

  \def\numtokens{60}
  \def\numsegments{6}
  \def\tokenwidth{1.5ex}
  \def\tokenheight{1.5ex}
  \pgfmathtruncatemacro{\segmentlength}{\numtokens / \numsegments}

  \coordinate (pos0) at (0, 0);
  \foreach[count=\j from 0] \segmentcolor in {%
    introcolor,
    charactercolor,
    challengecolor,
    strengthcolor,
    preventrycolor,
    entrycolor%
  }
  {
    \foreach \idx in {1,...,\segmentlength} {
      \pgfmathtruncatemacro{\i}{(\j * \segmentlength) + \idx}
      \pgfmathtruncatemacro{\prev}{\i - 1}
      \pgfmathtruncatemacro{\colorscale}{40 / \numtokens * \i}
      \pgfmathtruncatemacro{\tokenscale}{20 + random(40)}

      \node[Token, fill=tokencolor!\tokenscale!grey]
        (pos\i) at (pos\prev.east) {};

      \node[Token,
        yshift=-3 * \tokenheight,
        fill=poscolor1!\colorscale!poscolor2,
        minimum width=\tokenwidth]
        (tpos\i) at (pos\i.west) {};

      \ifthenelse{\idx=1 \OR \idx=2 \OR \idx=3}
      {
        \node[Token,
          fill=titlecolor!40!grey,
          yshift=-3 * \tokenheight,
          minimum width=\tokenwidth]
          (spos\i) at (tpos\i.west) {};
      }
      {
        \node[Token,
          fill=descriptioncolor!40!grey,
          yshift=-3 * \tokenheight,
          minimum width=\tokenwidth]
          (spos\i) at (tpos\i.west) {};
      }

      \node[Token,
        fill=\segmentcolor!40!grey,
        yshift=-3 * \tokenheight,
        minimum width=\tokenwidth]
        (s2pos\i) at (spos\i.west) {};
    };
  };

  \pgfmathtruncatemacro{\midpoint}{\numtokens / 2}
  \node[color=white, text=black,
    yshift=-1.5 * \tokenheight,
    minimum width=\tokenwidth,
    font={\huge}]
    (ppos\midpoint) at (pos\midpoint.east) {\bfseries +};

  \node[color=white, text=black,
    yshift=-1.5 * \tokenheight,
    minimum width=\tokenwidth,
    font={\huge}]
    (ppos\midpoint) at (tpos\midpoint.east) {\bfseries +};

  \node[color=white, text=black,
    yshift=-1.5 * \tokenheight,
    minimum width=\tokenwidth,
    font={\huge}]
    (ppos\midpoint) at (spos\midpoint.east) {\bfseries +};

  \node[Basic, anchor=south west]
    (tokenlabel)  at (pos1.north west) {Token Embeddings ($\bvec{e}$)};
  \node[Basic, anchor=south west]
    (positionlabel)  at (tpos1.north west) {Position Embeddings ($\bvec{p}$)};
  \node[Basic, anchor=south west]
    (segment1label)  at (spos1.north west) {Segment1 Embeddings ($\bvec{s_1}$)};
  \node[Basic, anchor=south west]
    (segment2label)  at (s2pos1.north west) {Segment2 Embeddings ($\bvec{s_2}$)};

  \foreach[count=\i] \len in {
    50, 50, 30, 20, 100, 250
  }
  {
    \pgfmathtruncatemacro{\startpt}{(\i - 1) * \segmentlength + 1}
    \pgfmathtruncatemacro{\endpt}{\startpt + \segmentlength - 1}

    \draw[|<->|,>=latex, font={\tiny}]
      ([yshift=-5pt] s2pos\startpt.south west) -- ([yshift=-5pt] s2pos\endpt.south east)
      node[anchor=north west, yshift=-5pt, xshift=1ex]
      (constraint\i) at (s2pos\startpt.south west)
      {Len $>=\len\ \vert\ $Pri $=\i$};
  }

  \node[Basic, yshift=-2pt, anchor=north]
    (constraintlabel) at (constraint\numsegments.south) {Constraint};

  \draw[decorate,decoration=brace]
    (constraint\numsegments.south east) -- (constraint\numsegments.south west);

   \node[Basic, yshift=-2.8em]
    (legend) at (s2pos1.west) {Segment Types:};

  \coordinate (legendlabel0) at (legend.east);
  \foreach[count=\i from 0] \segmentcolor / \segmentname / \loc / \height in {
    introcolor/intro/0/1.925pt,
    charactercolor/character/0/1.925pt,
    challengecolor/challenge card/1/0.925pt,
    strengthcolor/strength card/2/0.925pt,
    preventrycolor/prev entry/3/1.425pt,
    entrycolor/entry/4/1.425pt,
    titlecolor/title/5/1.925pt,
    descriptioncolor/description/6/0.925pt%
  }
  {
    \node[Basic, anchor=west]
      (legendlabel\i) at (legendlabel\loc.east)
      {\cbox[\segmentcolor]{\height}{\normalfont\footnotesize\segmentname}};
  };
\end{tikzpicture}
  \caption{An illustration of our segment embeddings and packing strategy. In
    addition to token and position embeddings, common to all Transformer
    models, we employ compositional segment embeddings for conditioning on
    story metadata (e.g., \cbox[strengthcolor]{2.75pt}{\textsc{deadly aim}} is
    the \cbox[titlecolor]{1.925pt}{title} of a
    \cbox[strengthcolor]{0.925pt}{strength card}). Each metadata segment has
    linear constraints with associated priorities (e.g., Len $>= 30\ \vert\ $ Pri
    $= 3$) for optimally packing tokens within the available space.}
  \label{fig:architecture}
\end{figure*}

We focus our modeling efforts on generating scene \emph{entries}, which are the
smallest units of each story, because we want to evaluate the generated text on
the \storium\ platform within a machine-in-the-loop framework.\footnote{Our
dataset also enables modeling high-level decisions made by the narrator, such as
challenge sequencing; we leave this for future work.} Our method relies on
fine-tuning a pretrained language model (GPT-2) on the \storium\ dataset using
segment embeddings to differentiate each type of context. While GPT-2 has
successfully been used as a state-of-the-art model for story
generation~\cite{mao-etal-2019-improving,guan-etal-2020-knowledge}, one crucial
challenge is the length of the contexts: each entry in a story can condition on
any narrative element that comes before it (e.g., previous entries, scenes,
challenges). Thus, the number of context tokens quickly grows larger than what
is feasible to fit in GPU memory. Another challenge lies in how to properly
tune hyperparameters in a machine-in-the-loop setting, as it is infeasible to
obtain human judgments for a huge number of configurations. The rest of this
section fully specifies our model, a token-packing strategy to optimize use of
the input context, and preliminary user-facing experiments that helped us
decide on our final model hyperparameters.

\subsection{Model Specification}
We fine-tune the GPT-2 medium-sized (355M parameters) language model
\cite{radfordGPT2} for story generation, as it has been shown to generate
coherent long-form prose. Before fine-tuning, we need to account for the
complexity of \storium\ contexts: each scene consists of multiple entries, each
of which may reference a different number of semi-structured cards (e.g., both
the \cbox[strengthcolor]{2.75pt}{\textsc{deadly aim}} \cbox[strengthcolor]{0.925pt}{strength card} and the
\cbox[charactercolor]{1.925pt}{\textsc{adira makarova}} \cbox[charactercolor]{0.925pt}{character} in
\figureref{example_cards} contain a title and description). To handle the
compositional and semi-structured nature of the scenes and cards, we allow each
input token to condition on an arbitrary number of \emph{segment embeddings}
\cite{Wolf2019TransferTransfoAT} (\figureref{architecture}). Concretely, we
augment the token vocabulary $V$ of GPT-2 with a segment vocabulary $S$ for
delineating each segment. The final embedding vector $\bvec{e}_i$ at position
$i$ is computed by summing the token embedding $\bvec{v}_i$ with the positional
embedding $\bvec{p}_i$ and the corresponding set of $n$ segment embeddings
$\{\bvec{s}_{i_1},\ldots,\bvec{s}_{i_n}\}$:

\begin{equation}
  e_i = p_i + v_i + \sum_{m=1}^ns_{i_m}
\end{equation}

During training, a single input instance to our models contains the text of the
current entry, its associated challenge, card metadata, as well as the
current character's biography and the scene's introductory text
(\figureref{example_cards}). Our final model also includes the text of the
immediately preceding story entry,\footnote{If the preceding entry
is not written by the current character, we also include the current
character's last entry.} which improves human and automatic evaluation scores (\tableref{exploratory}).
At test time, we provide only the story context and
autoregressively sample a scene entry. 

\subsubsection{Context packing}
The average story in our dataset has over 19K tokens broken up into 78 scene
entries, which is much longer than GPT-2's maximum sequence length of 1024
tokens. We thus face the challenge of how best to optimize our usage of the
limited input space, which is made more difficult by the many different types of
input context (e.g., entries, characters, challenges) within
\storium. Na\"ively reserving a fixed number of tokens per context type wastes
significant space, as the number and length of metadata instances varies
considerably per entry. For example, some scene entries do not make use of cards
(\tableref{dataset-condensed}), while others reference multiple cards.

Our solution applies the Cassowary algorithm \cite{Badros2001TheCL}, well-known
for arranging UI elements in Apple's iOS, to pack the input tokens more
efficiently. Cassowary allows for efficiently solving linear equality and
inequality constraints incrementally, using a dual simplex based method. We
define a set of linear constraints on the size of each metadata segment (e.g.,
include at least 250 tokens from an entry when possible), and Cassowary's
solver produces an optimal arrangement of context tokens with respect to these
constraints (\figureref{architecture}). Compared to na\"ively packing
tokens into fixed length segments, Cassowary allows us to vary the minimum and
maximum bounds on segments, as well as collapse missing segments. This
flexibility results in increased human and automatic evaluation scores
(\tableref{exploratory}).

\subsection{Hyperparameter Selection}
\label{sec:hyperparams}
Before launching our full machine-in-the-loop evaluation, we conduct preliminary
experiments on the \storium\ platform to validate our design choices. Since we
want \emph{real} users on \storium\ to enjoy interacting with the generated
text, we want to avoid alienating them with poorly performing models. We
 measure the impact of (1) including history information from the
immediately preceding entry in the story, and (2) using Cassowary to densely
pack the context. In total, we fine-tune four models on the Cartesian product
of these complementary modeling ideas, keeping all other hyperparameters
constant, and deploy these models to \storium.

The results (\tableref{exploratory}) highlight the importance of both modeling
choices: after including more story history and applying the Cassowary solver,
validation perplexity decreases while \storium\ user ratings of fluency,
coherence, relevance, and likability all increase. This motivates us to use
only the best-performing model for the full-scale evaluation. Additionally,
user feedback from these experiments suggested that we generate \emph{shorter}
entries, as longer ones frequently devolved into unrelated and incoherent
sentences. Thus, for our final experiments detailed in the next section, we
also truncate model outputs to a maximum of four sentences. 

\begin{table}[t]
  \centering
  \small
  \begin{tabular}{c|c|cccc|c|c}
  Cas & His & F & C & L & R & Ppl & Jdg \\
  \hline
  & & 3.4 & 2.9 & 3.8 & 2.3 & 25.1 & 90 \\
  & \checkmark & 3.1 & 2.7 & 3.9 & 2.3 & 22.4 & 77 \\
  \checkmark & & 3.6 & 2.8 & 3.6 & 2.4 & 22.9 & 62 \\
  \checkmark & \checkmark & \bf{3.7} & \bf{3.2} & \bf{4.1} & \bf{2.7} & \bf{21.0} & 85 \\
\end{tabular}

  \caption{Exploratory experiments indicate optimally packing tokens using
    Cassowary (Cas), and including more history (His) is key to achieving low
    perplexity (Ppl), along with high fluency (F), coherence (C), likability
    (L), and relevance (R) based on a number of user judgments (Jdg).}
  \label{table:exploratory}
\end{table}

\section{A Machine-in-the-Loop Evaluation Platform}
\label{sec:methodology}
The inadequacies of existing human and automatic evaluation methods are a
major roadblock for story generation research. Automatic evaluations
correlate weakly with human judgments~\citep{sagarkar2018quality}, and these
judgments are obtained from crowd workers who are not invested in the
narratives they are assessing. These concerns are magnified with \storium, as
the story contexts are far too long for crowd workers to reliably evaluate
(\sectionref{mturk-analysis}). In this section, we propose an improved
evaluation methodology by directly integrating our models onto the \storium\
platform. This allows story \emph{authors} to query a
machine~\citep{Clark2018CreativeWW} for suggestions during the process of
writing their own stories. We develop a new evaluation metric,
\storymetricfull\ (\storymetric), computed on top of the \emph{edits} that
\storium\ users make to generated entries. Finally, we provide experimental
results that compare two configurations of our best model from
\sectionref{hyperparams}.

\subsection{Evaluation Lifecycle}
To evaluate generated stories, we develop a dedicated web
service for serving model outputs to the \storium\ platform. \storium\ users
simply press a button on the user interface to obtain a generated scene entry
conditioned on the story context.  Users can then \highlightinline[green]{add}
new text while \highlightinline[red]{deleting} any of the generated text that
they wish (\figureref{example_cards}). When users publish their edited
entry, they are also asked to evaluate the generated text on a 5-point Likert
scale\footnote{They also provide optional freeform comments on generated text;
we leave analysis of the comments to future work.} with respect to \textit{relevance} (fit with the current story),
\textit{fluency} (judgment of grammaticality), \textit{coherence} (logical
ordering of sentences), and \textit{likability} (subjective assessment of
enjoyability). This process allows experts (\storium\ authors) to evaluate
generated stories, which is a substantial improvement over prior evaluation
efforts. We make our evaluation platform publicly accessible for researchers to
develop and integrate their own models. Our framework makes adding a new model
using any Python-based deep learning framework very easy, requiring
implementation of only four methods: \texttt{startup}, \texttt{shutdown},
\texttt{preprocess}, and \texttt{generate}.

\subsection{A Metric Over User Edits}
\label{sec:metrics}
Intuitively, the amount of generated text that a user preserves in their final
published entry clearly indicates the usefulness of the generated text. We
quantify this by developing \storymetricfull\ (\storymetric), inspired by
the longest common subsequence (LCS) variant of \rouge~\cite{Lin2004ROUGE},
applied to user edits. Given a generated entry $X$ and the final published
entry $Y$, we compute $\storymetric(X,Y)=\frac{|\textsc{match}(X,Y)|}{|X|}$,
where $\textsc{match}(X,Y)$ considers \emph{contiguous substrings} with at least
one non-stopword as \highlightinline[yellow]{matches} (see
\figureref{example_cards} for an example and \appendixref{metric} for a more
thorough treatment). We do not use \rougel\ because vanilla LCS typically
favors subsequences of unigram matches (often stopwords) over longer contiguous
n-gram matches. In our \storium\ setting, users preserving n-grams or full
sentences is a clear indication that the generated text was useful.

\section{Analysis}
\label{sec:analysis}

Compared to existing work on story generation, the main novelty of our \storium\
evaluation platform is that it enables authors to interact directly with
model-generated text through their edits. In this section, we conduct
experiments on our platform and analyze the edits by examining the correlation
of \storymetric\ to Likert scores. We explore linguistic properties of
text that users preserve and also conduct a crowdsourced evaluation on Amazon
Mechanical Turk that demonstrates its unsuitability for this task. Finally, we
qualitatively describe feedback obtained from interviews with ten \storium\
users who engaged with our models, which provides a roadmap for future work.

\begin{table}[tb]
  \centering
  \small
  \def\nsig{\makebox[0pt][l]{\textsuperscript{$\dagger$}}}
\begin{tabular}{p{1.5ex}r|c|c|c|c||c}
  & & Lik & Flu & Coh & \storymetric & Rating\\
  \hline

  Rel & top-$k$ & 0.51
    & 0.28
    & 0.55
    & 0.51
    & \textbf{2.55} \\

    & nucleus & 0.53
    & 0.40
    & 0.57
    & 0.39
    & 2.47 \\
  \hline

  Lik & top-$k$ & ---
    & 0.28
    & 0.35
    & 0.34
    & \textbf{3.32} \\

    & nucleus & ---
    & 0.38
    & 0.55
    & 0.35
    & 3.21 \\
  \hline

  Flu & top-$k$ & ---
    & ---
    & 0.54
    & 0.13\nsig
    & \textbf{3.96} \\

    & nucleus & ---
    & ---
    & 0.61
    & 0.23
    & 3.76 \\
  \hline

  Coh & top-$k$ & ---
    & ---
    & ---
    & 0.25
    & \textbf{3.41} \\

    & nucleus & ---
    & ---
    & ---
    & 0.36
    & 2.96 \\
  \hline

  \storymetric\ & top-$k$ & ---
    & ---
    & ---
    & ---
    & \textbf{15.63} \\

    & nucleus & ---
    & ---
    & ---
    & ---
    & 9.86 \\

  \hline
\end{tabular}

  \caption{Despite its low rating, relevance is clearly important as indicated
  by the moderately strong Pearson's $r$ correlations (first four columns)
  with \storymetric\ and the remaining human judgments. All correlations are
  significant ($p<0.01$), except those indicated by $\dagger$ ($p>0.05$).}
  \label{table:baseline-correlations}
\end{table}

\paragraph{Top-$k$ vs. nucleus sampling:}
\label{sec:experiments}

Using our platform (\sectionref{methodology}), we evaluate our best model
(\tableref{exploratory}) with two different decoding strategies: (1)
top-$k$ sampling~\citep{fan2018hierarchical} with $k=40$, and (2) nucleus
sampling~\citep{Holtzman2020TheCC} with $p=0.9$.\footnote{We use a temperature
of 0.9, a repetition penalty \cite{Keskar2019CTRLAC} of 1.2, and an analogous
length penalty that dynamically penalizes producing the end of sequence token
inversely proportionally to a desired length $l_d$.} The sampling parameters, such as the $k$ in
top-$k$ sampling, can significantly affect output quality of story generation
models~\citep{see2019massively}, so we choose values that worked well in prior
work~\citep{qin2019counterfactual}.\footnote{\label{footnote:sampling-caveat}It
is possible that a better set of sampling hyperparameters exists, which we
leave to future work.} 

Interestingly, while~\citet{Holtzman2020TheCC} show that nucleus sampling
improves over top-$k$ sampling on measures like repetition, \storium\
users clearly prefer the top-$k$ variant across all categories (last column of
\tableref{baseline-correlations}). We collect roughly 200 feedback ratings and
175 edits for each model over a span of three months beginning in late February
2020. We discover that both configurations score best on \emph{fluency} and
worst on \emph{relevance}. This is unsurprising as (1) GPT-2 is known to
produce fluent text and (2) the complex and lengthy \storium\ data is a
challenge for limited-context models. Finally, \storymetric\ scores are
generally low (15.6 for top-$k$ vs. 9.9 for nucleus sampling),
indicating that users delete most of the current model's generated text. This
result demonstrates that story generation models still have a long way to go.\footnote{\label{footnote:supplementary-html}See the supplementary HTML
for an export of all results (including generated text and edits) used for this paper.}

\paragraph{\storymetric\ correlates with human judgments:}
\label{sec:correlations}
A natural question is whether our \storymetric\ metric correlates with
judgments of fluency, coherence, relevance, and likability.
\tableref{baseline-correlations} shows that for the top-$k$ configuration,
relevance has a significantly higher correlation (Pearson's $r$) with
\storymetric\ than the other properties. In other words, users are most likely
to preserve generated text when it is relevant to the overall story. Fluency
correlates only weakly with \storymetric, which makes sense as most generated
entries are fluent due to GPT-2's pretraining. Finally, nucleus sampling
exhibits lower correlation for relevance, but higher correlation for the other
three properties, possibly due to its lower average scores for these properties
(see \appendixref{metric} for a comparison of \storymetric\ to \rouge-based
metrics).\footnoteref{supplementary-html}

\paragraph{Linguistic properties of preserved text:}
\label{sec:linguistic-analysis}
Knowing that users delete most of the generated text, we instead explore the
linguistic commonalities of the preserved text. We run spaCy part-of-speech
tagging and named entity recognition~\citep{spacy2} over the edited entries.
Strikingly, 29.5\% of generated proper nouns are preserved in the edited text,
compared to only 13.5\% for all other POS tags. A major confound is that our
model could unfairly receive credit for simply copying character names from the
input context, as users are likely to write about these characters anyway. To
measure the extent of this effect, we match all generated named entities that
users preserve to predefined character lists from each story, and discover that
63\% of generated entities already exist within the story context. The
remaining 37\% of entities are often completely new character names. User
interviews also suggest that this ability to generate new names is a useful
feature.

\begin{table}[tb]
  \centering
  \small
  \begin{tabular}{lcccc}
  & \multicolumn{2}{c}{Top-$k$} & \multicolumn{2}{c}{Nucleus} \\
  First Run & Rating & $\kappa$ & Rating & $\kappa$\\
  \hline
  \multicolumn{1}{l|}{Fluency}   & \textbf{3.59} & 0.17 & 3.47 & 0.11 \\
  \multicolumn{1}{l|}{Coherence} & \textbf{3.50} & 0.10 & 3.44 & 0.20 \\
  \multicolumn{1}{l|}{Likability}& \textbf{3.27} & 0.07 & 3.22 & 0.11 \\
  \multicolumn{1}{l|}{Relevance} & \textbf{3.32} & 0.09 & 3.27 & 0.13 \\
  \hline
  &  &  &  &  \\
  & \multicolumn{2}{c}{Top-$k$} & \multicolumn{2}{c}{Nucleus} \\
  Second Run & Rating & $\kappa$ & Rating & $\kappa$\\
  \hline
  \multicolumn{1}{l|}{Fluency}   & \textbf{4.01} & 0.46 & 3.77 & 0.33 \\
  \multicolumn{1}{l|}{Coherence} & \textbf{3.63} & 0.27 & 3.38 & 0.23 \\
  \multicolumn{1}{l|}{Likability}& \textbf{3.28} & 0.12 & 3.06 & 0.16 \\
  \hline
\end{tabular}

  \caption{Despite our best efforts, our first crowd sourced judgments show low
    agreement ($\kappa$) on open-ended story generation. Our second run, which
    removes context, thus excluding relevance judgments, greatly increases
    agreement for fluency and coherence.}
  \label{table:crowd-workers}
\end{table}

\paragraph{Crowdsourced evaluation is unreliable:}
\label{sec:mturk-analysis}
Thus far, we have argued for our evaluation platform by claiming that
crowdsourced methods are unsuitable for evaluating stories with complex and
lengthy contexts. Here, we measure fluency, coherence, relevance, and
likability of our generated entries with a crowdsourced Amazon Mechanical Turk
task, to see if the results correspond to \storium\ user ratings. Designing this
crowdsourced task is difficult, as we cannot show crowd workers
the entire story context due to its length; we thus decide to show the same
inputs that the model receives (\sectionref{model}). We collect ratings of
100 examples per model, with three judgments per example.\footnote{We limit
annotations to crowd workers living in the US and the UK, with over 1000
completed annotations and a 99\% approval. We pay \$0.50 per annotation, by
assuming 2 minutes per annotation, for an effective hourly rate of \$15.}

\tableref{crowd-workers} (top) shows that workers have very low
agreement (Fleiss' $\kappa$) for all properties, including even fluency. An
analysis of the median task completion time\footnote{Mechanical Turk automatically
reports a \textit{WorkTimeInSeconds} field for each annotation, which is ten
minutes on average for our task --- more than enough time to
read and assess the generated entry and associated context. Sadly, this
interval is misleading. Analyzing the median time between submits, we see
workers accept multiple concurrent tasks, wait a few minutes, then submit each
annotation in quick succession, thus inflating the \textit{WorkTimeInSeconds}
interval.} reveals most workers did not actually read the context. We run a
second experiment, showing only the generated text (no context), and remove the
relevance rating. \tableref{crowd-workers} (bottom) shows this improves
agreement (\tableref{crowd-workers}), and that the average fluency scores align
closely with those from \storium\ users. Overall, our struggle to obtain
quality judgments from Mechanical Turk further validates our platform:
\storium\ provides free expert judgments from people invested in storytelling.

\paragraph{Feedback from user interviews:}
\label{sec:interviews}
To better understand the strengths and weaknesses of our current
model, we conduct semi-structured interviews with ten \storium\ users. Most were
surprised with the overall fluency of our models. This partly
explains the low correlation of fluency with \storymetric. Relevance was
mentioned by 9 out of 10 users as the number one area of improvement for our
model, confirming our experimental results
(\tableref{baseline-correlations}). Four users called out the model's
tendency to fabricate facts and introduce new characters. Despite these
concerns, three users explicitly stated the model inspired them to write or
found portions of the generated text useful, though mostly as a source for
character and place names (supporting the linguistic analysis in
\sectionref{linguistic-analysis}). Finally, some users considered the system a
curiosity and decided to write stories using only generated text (without
edits).\footnote{These AI-guided narratives are prevalent enough that we
manually exclude these games from our experiments as they artificially
increase the automatic metrics.}

\section{Related Work}
\label{sec:relwork}

Our work builds on prior research in computational modeling for story
generation. Early narrative prose generation systems
\cite{Meehan1977TALESPINAI,Callaway2001NarrativePG,Riedl2004AnIP} relied on
graph-based planning formalisms and custom rules to structure their narratives,
while story graphs have been used for interactive
storytelling~\cite{Riedl2013InteractiveNA}. More recent work uses deep learning
to generate stories by training neural models with limited
context~\cite{peng2018towards,fan2018hierarchical,GoldfarbTarrant2019PlanWA}
and structured knowledge, either
external~\cite{mao-etal-2019-improving,guan-etal-2020-knowledge,goldfarb2020content}
or derived ~\cite{yao2019plan,fan2019strategies}. Compared to the datasets
studied in those works, our \storium\ dataset contains much longer stories with
built-in structural annotations written in natural language in the form of
cards (\tableref{dataset-condensed}).

Our work connects more closely to existing machine-in-the-loop storytelling
work \cite{roemmele2015creative,Samuel2016WritingBuddy,Clark2018CreativeWW}, in
which systems work in concert with users to collaboratively author a narrative.
Much like the Creative Help platform of~\citet{roemmele2015creative}, we
provide writing assistance by interactively generating continuations of
\storium\ stories. We improve over~\citet{roemmele2015creative} by evaluating a
trained model (instead of a retrieval-based approach) with a large user
population.

Finally, our \storium\ evaluation takes a different approach to prior research
that measures the quality of generated stories. \citet{sagarkar2018quality}
train an automatic scorer on human annotations of overall story quality,
relevance, and interestingness based on evaluation criteria from
\citep{McIntyre2009LearningTT}.~\citet{see2019massively} consider a number of
\textit{diversity} related measures for automated evaluation of story
generation systems by focusing on the GPT-2 small model, noting that
\textit{quality} assessments are still best measured through human evaluation.

\section*{Limitations}
\label{sec:limitations}

Evaluating on the \storium\ platform enables researchers to receive high-quality judgements on the outputs of their story generation models. These judgements are made possible by the significant time and effort spent by real authors on
crafting their narratives, as their incentives are substantially different than those of crowdsourced workers. The amount of author effort involved in evaluation, when combined with the relatively small size of the \storium\ community,
can cause evaluation to take a considerable amount of time (i.e., to collect hundreds of judgements) as evidenced in our
analysis (Section~\ref{sec:analysis}). Thus, our platform is not currently suitable for ``instant'' evaluation of generated stories. Furthermore, as the evaluation platform
is specifically deployed on \storium, it cannot be trivially used to evaluate models trained on other story generation
datasets, as users of the website are mainly invested in writing narratives that follow the \storium\ format.

\section{Conclusion}
\label{sec:conclusions}

We introduce the \storium\ dataset and evaluation platform for
machine-in-the-loop story generation, built from an online collaborative
storytelling community. \storium\ contains 6K long stories annotated with
structural metadata useful for conditioning language models. Importantly,
\emph{real \storium\ authors} evaluate model outputs by adding and
removing text to create their own stories. We devise a metric on top of their
edits that correlates strongly with judgments of the \emph{relevance} of the
generated text, which user interviews suggest is the most important
area for improvement moving forward. Our dataset and evaluation platform will
be made publicly available to spur progress into story generation. 

\ifaclfinal
\section*{Author Contributions}
\label{sec:contributions}

Dataset Analysis: Akoury, Wang \\
Generation Model: Akoury, Wang \\
Evaluation Platform: Akoury, Whiting, Hood \\
Research Guidance: Iyyer, Peng

\section*{Acknowledgements}
\label{sec:acknowledge}

We thank the wonderful \storium\ users for actively using our story generation
models and generously providing their time to be interviewed. We also thank the
amazing UMass NLP community for thoughtful insights on our paper and helping to
validate whether structural metadata influences story text on \storium. Akoury
and Iyyer were supported during this project by a research gift from Genpact.
Peng was supported in part by the CwC program under Contract W911NF-15-1-0543
with the US Defense Advanced Research Projects Agency (DARPA).

\fi

\bibliography{2020_emnlp_storyteller}
\bibliographystyle{style/acl_natbib}

\newpage
~
\newpage

\setcounter{table}{0} \renewcommand{\thetable}{A\arabic{table}}
\setcounter{figure}{0} \renewcommand{\thefigure}{A\arabic{figure}}
\setcounter{footnote}{0} \renewcommand{\thefootnote}{\arabic{footnote}}

\appendix
%
\makeatletter
\providecommand*{\input@path}{}
\edef\input@path{{./}\input@path}
\makeatother

\section*{Appendix}

\section{Additional Dataset Statistics}
\label{sec:storium_stats}

As our dataset derives from a collaborative storytelling game that is highly
compositional by nature, it is difficult to concisely capture the full scope of
the data within the main body. Here we highlight the full results of our small
scale annotation that indicates cards influence the scene entry text.

\begin{table}[H]
  \centering
  \small
  \def\callout{\makebox[0pt][l]{\textsuperscript{$\dagger$}}}
\begin{tabular}{|lr|}
\hline
  Total Annotations&248 \\
  Valid Entries\callout&235 \\
  Card Influences Entry&182 \\
  Entry Addresses Challenge&189 \\
  Card Influence $\cap$ Challenge Addressed&151 \\ \hline
\end{tabular}

  \caption{We ask annotators to determine how frequently cards influence an
  entry, and if the entry addresses the challenge. $\dagger$Annotators were
  asked to flag stories not written in English or otherwise could not be
  understood.}
  \label{table:card-influence}
\end{table}

Additionally, there are many small details which are important distinctions in
the game, but may not require separate modeling for generating a scene entry.
For example, there is a distinction between regular cards, which have a fixed
title and description provided by the narrator; versus wild cards, which allow
individual characters to write their own title and description. For the sake of
completeness, we provide \tableref{storium_stats} to help further explore the
depths of this unique dataset. The following
histograms\footnote{\label{footnote:mean-stddev} These histograms provide context
for the meaning of \textbf{Mean} and \textbf{Std Dev} for
\tableref{storium_stats}.} further break down the data in
\tableref{storium_stats}, clearly demonstrating the long tail distributions
indicative of user generated stories:

\begin{table*}[tb]
  \centering
  \begin{tabular}{|l|c|c|c|}
\hline
\bf Feature & \bf Total & \bf Mean\textsuperscript{\ref{footnote:mean-stddev}} & \bf Std Dev\textsuperscript{\ref{footnote:mean-stddev}} \\
\hline
Stories&5,743&---&--- \\
\hline
Completed Stories&586&---&--- \\
\hline
Users&30,119&---&--- \\
\hline
Characters Created&27,462&4.78&3.04 \\
\hline
Characters Played&25,955&4.52&3.04 \\
\hline
Total Played Roles&31,698&---&--- \\
\hline
Scenes&25,092&4.37&6.96 \\
\hline
Scene Entries&448,264&17.86&19.37 \\
\hline
Cards Created/Edited&318,692&55.49&56.88 \\
\hline
Total Played Cards by Users&232,596&40.50&70.94 \\
\hline
Played Cards Created/Edited by Users&204,698&35.64&67.65 \\
\hline
Location Cards Played by Narrators&16,887&0.67&0.47 \\
\hline
Challenge Cards Played by Narrators&61,223&0.47&1.05 \\
\hline
Cards Played by Characters&149,014&0.47&0.63 \\
\hline
Wild Cards Played by Characters&31,465&0.10&0.30 \\
\hline
Regular Cards Played by Characters&117,549&0.37&0.58 \\
\hline
Stories Played Without Cards&736&---&--- \\
\hline
Tokens in Character Descriptions&7,155,548&260.56&286.78 \\
\hline
Tokens in Scene Entries&110,772,426&247.11&307.13 \\
\hline
Tokens in Played Location Cards&438,044&0.98&6.14 \\
\hline
Tokens in Played Challenge Cards&3,837,860&24.84&15.30 \\
\hline
Tokens in Played Regular Cards&3,053,152&25.08&15.78 \\
\hline
Tokens in Played Wild Cards&784,708&23.96&13.32 \\
\hline
Unique Tokens&424,768&---&--- \\
\hline
\end{tabular}

  \caption{A small look at the highly compositional nature of our dataset.}
  \label{table:storium_stats}
\end{table*}

\resizebox{\columnwidth}{!}{
\begingroup
  \makeatletter
  \providecommand\color[2][]{%
    \GenericError{(gnuplot) \space\space\space\@spaces}{%
      Package color not loaded in conjunction with
      terminal option `colourtext'%
    }{See the gnuplot documentation for explanation.%
    }{Either use 'blacktext' in gnuplot or load the package
      color.sty in LaTeX.}%
    \renewcommand\color[2][]{}%
  }%
  \providecommand\includegraphics[2][]{%
    \GenericError{(gnuplot) \space\space\space\@spaces}{%
      Package graphicx or graphics not loaded%
    }{See the gnuplot documentation for explanation.%
    }{The gnuplot epslatex terminal needs graphicx.sty or graphics.sty.}%
    \renewcommand\includegraphics[2][]{}%
  }%
  \providecommand\rotatebox[2]{#2}%
  \@ifundefined{ifGPcolor}{%
    \newif\ifGPcolor
    \GPcolortrue
  }{}%
  \@ifundefined{ifGPblacktext}{%
    \newif\ifGPblacktext
    \GPblacktextfalse
  }{}%
  \let\gplgaddtomacro\g@addto@macro
  \gdef\gplbacktext{}%
  \gdef\gplfronttext{}%
  \makeatother
  \ifGPblacktext
    \def\colorrgb#1{}%
    \def\colorgray#1{}%
  \else
    \ifGPcolor
      \def\colorrgb#1{\color[rgb]{#1}}%
      \def\colorgray#1{\color[gray]{#1}}%
      \expandafter\def\csname LTw\endcsname{\color{white}}%
      \expandafter\def\csname LTb\endcsname{\color{black}}%
      \expandafter\def\csname LTa\endcsname{\color{black}}%
      \expandafter\def\csname LT0\endcsname{\color[rgb]{1,0,0}}%
      \expandafter\def\csname LT1\endcsname{\color[rgb]{0,1,0}}%
      \expandafter\def\csname LT2\endcsname{\color[rgb]{0,0,1}}%
      \expandafter\def\csname LT3\endcsname{\color[rgb]{1,0,1}}%
      \expandafter\def\csname LT4\endcsname{\color[rgb]{0,1,1}}%
      \expandafter\def\csname LT5\endcsname{\color[rgb]{1,1,0}}%
      \expandafter\def\csname LT6\endcsname{\color[rgb]{0,0,0}}%
      \expandafter\def\csname LT7\endcsname{\color[rgb]{1,0.3,0}}%
      \expandafter\def\csname LT8\endcsname{\color[rgb]{0.5,0.5,0.5}}%
    \else
      \def\colorrgb#1{\color{black}}%
      \def\colorgray#1{\color[gray]{#1}}%
      \expandafter\def\csname LTw\endcsname{\color{white}}%
      \expandafter\def\csname LTb\endcsname{\color{black}}%
      \expandafter\def\csname LTa\endcsname{\color{black}}%
      \expandafter\def\csname LT0\endcsname{\color{black}}%
      \expandafter\def\csname LT1\endcsname{\color{black}}%
      \expandafter\def\csname LT2\endcsname{\color{black}}%
      \expandafter\def\csname LT3\endcsname{\color{black}}%
      \expandafter\def\csname LT4\endcsname{\color{black}}%
      \expandafter\def\csname LT5\endcsname{\color{black}}%
      \expandafter\def\csname LT6\endcsname{\color{black}}%
      \expandafter\def\csname LT7\endcsname{\color{black}}%
      \expandafter\def\csname LT8\endcsname{\color{black}}%
    \fi
  \fi
    \setlength{\unitlength}{0.0500bp}%
    \ifx\gptboxheight\undefined%
      \newlength{\gptboxheight}%
      \newlength{\gptboxwidth}%
      \newsavebox{\gptboxtext}%
    \fi%
    \setlength{\fboxrule}{0.5pt}%
    \setlength{\fboxsep}{1pt}%
\begin{picture}(4680.00,3276.00)%
    \gplgaddtomacro\gplbacktext{%
      \colorrgb{0.00,0.00,0.00}
      \put(1078,811){\makebox(0,0)[r]{\strut{}$1$}}%
      \colorrgb{0.00,0.00,0.00}
      \put(1078,1132){\makebox(0,0)[r]{\strut{}$4$}}%
      \colorrgb{0.00,0.00,0.00}
      \put(1078,1452){\makebox(0,0)[r]{\strut{}$16$}}%
      \colorrgb{0.00,0.00,0.00}
      \put(1078,1773){\makebox(0,0)[r]{\strut{}$64$}}%
      \colorrgb{0.00,0.00,0.00}
      \put(1078,2093){\makebox(0,0)[r]{\strut{}$256$}}%
      \colorrgb{0.00,0.00,0.00}
      \put(1078,2414){\makebox(0,0)[r]{\strut{}$1024$}}%
      \colorrgb{0.00,0.00,0.00}
      \put(1078,2734){\makebox(0,0)[r]{\strut{}$4096$}}%
      \colorrgb{0.00,0.00,0.00}
      \put(1078,3055){\makebox(0,0)[r]{\strut{}$16384$}}%
      \colorrgb{0.00,0.00,0.00}
      \put(1257,632){\rotatebox{-45}{\makebox(0,0)[l]{\strut{}$0$}}}%
      \colorrgb{0.00,0.00,0.00}
      \put(1761,632){\rotatebox{-45}{\makebox(0,0)[l]{\strut{}$50$}}}%
      \colorrgb{0.00,0.00,0.00}
      \put(2266,632){\rotatebox{-45}{\makebox(0,0)[l]{\strut{}$100$}}}%
      \colorrgb{0.00,0.00,0.00}
      \put(2770,632){\rotatebox{-45}{\makebox(0,0)[l]{\strut{}$150$}}}%
      \colorrgb{0.00,0.00,0.00}
      \put(3274,632){\rotatebox{-45}{\makebox(0,0)[l]{\strut{}$200$}}}%
      \colorrgb{0.00,0.00,0.00}
      \put(3779,632){\rotatebox{-45}{\makebox(0,0)[l]{\strut{}$250$}}}%
      \colorrgb{0.00,0.00,0.00}
      \put(4283,632){\rotatebox{-45}{\makebox(0,0)[l]{\strut{}$300$}}}%
    }%
    \gplgaddtomacro\gplfronttext{%
      \csname LTb\endcsname
      \put(198,1933){\rotatebox{-270}{\makebox(0,0){\strut{}Count}}}%
      \put(2770,154){\makebox(0,0){\strut{}Entries per Scene}}%
      \csname LTb\endcsname
      \put(4132,2775){\makebox(0,0)[r]{\strut{}total=448264}}%
      \put(4132,2606){\makebox(0,0)[r]{\strut{} mean=17.86}}%
      \put(4132,2438){\makebox(0,0)[r]{\strut{} std=19.37}}%
      \put(4132,2270){\makebox(0,0)[r]{\strut{}bin width=10}}%
    }%
    \gplbacktext
    \put(0,0){\includegraphics{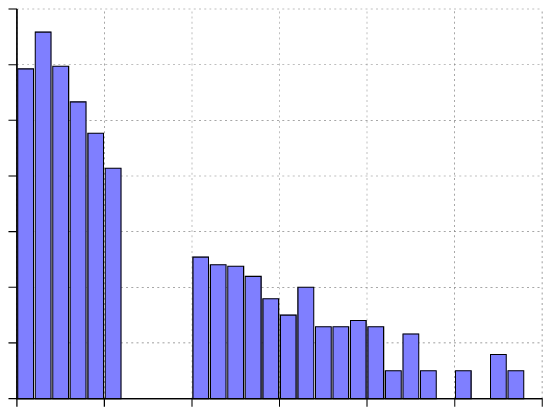}}%
    \gplfronttext
  \end{picture}%
\endgroup
}
\resizebox{\columnwidth}{!}{
\begingroup
  \makeatletter
  \providecommand\color[2][]{%
    \GenericError{(gnuplot) \space\space\space\@spaces}{%
      Package color not loaded in conjunction with
      terminal option `colourtext'%
    }{See the gnuplot documentation for explanation.%
    }{Either use 'blacktext' in gnuplot or load the package
      color.sty in LaTeX.}%
    \renewcommand\color[2][]{}%
  }%
  \providecommand\includegraphics[2][]{%
    \GenericError{(gnuplot) \space\space\space\@spaces}{%
      Package graphicx or graphics not loaded%
    }{See the gnuplot documentation for explanation.%
    }{The gnuplot epslatex terminal needs graphicx.sty or graphics.sty.}%
    \renewcommand\includegraphics[2][]{}%
  }%
  \providecommand\rotatebox[2]{#2}%
  \@ifundefined{ifGPcolor}{%
    \newif\ifGPcolor
    \GPcolortrue
  }{}%
  \@ifundefined{ifGPblacktext}{%
    \newif\ifGPblacktext
    \GPblacktextfalse
  }{}%
  \let\gplgaddtomacro\g@addto@macro
  \gdef\gplbacktext{}%
  \gdef\gplfronttext{}%
  \makeatother
  \ifGPblacktext
    \def\colorrgb#1{}%
    \def\colorgray#1{}%
  \else
    \ifGPcolor
      \def\colorrgb#1{\color[rgb]{#1}}%
      \def\colorgray#1{\color[gray]{#1}}%
      \expandafter\def\csname LTw\endcsname{\color{white}}%
      \expandafter\def\csname LTb\endcsname{\color{black}}%
      \expandafter\def\csname LTa\endcsname{\color{black}}%
      \expandafter\def\csname LT0\endcsname{\color[rgb]{1,0,0}}%
      \expandafter\def\csname LT1\endcsname{\color[rgb]{0,1,0}}%
      \expandafter\def\csname LT2\endcsname{\color[rgb]{0,0,1}}%
      \expandafter\def\csname LT3\endcsname{\color[rgb]{1,0,1}}%
      \expandafter\def\csname LT4\endcsname{\color[rgb]{0,1,1}}%
      \expandafter\def\csname LT5\endcsname{\color[rgb]{1,1,0}}%
      \expandafter\def\csname LT6\endcsname{\color[rgb]{0,0,0}}%
      \expandafter\def\csname LT7\endcsname{\color[rgb]{1,0.3,0}}%
      \expandafter\def\csname LT8\endcsname{\color[rgb]{0.5,0.5,0.5}}%
    \else
      \def\colorrgb#1{\color{black}}%
      \def\colorgray#1{\color[gray]{#1}}%
      \expandafter\def\csname LTw\endcsname{\color{white}}%
      \expandafter\def\csname LTb\endcsname{\color{black}}%
      \expandafter\def\csname LTa\endcsname{\color{black}}%
      \expandafter\def\csname LT0\endcsname{\color{black}}%
      \expandafter\def\csname LT1\endcsname{\color{black}}%
      \expandafter\def\csname LT2\endcsname{\color{black}}%
      \expandafter\def\csname LT3\endcsname{\color{black}}%
      \expandafter\def\csname LT4\endcsname{\color{black}}%
      \expandafter\def\csname LT5\endcsname{\color{black}}%
      \expandafter\def\csname LT6\endcsname{\color{black}}%
      \expandafter\def\csname LT7\endcsname{\color{black}}%
      \expandafter\def\csname LT8\endcsname{\color{black}}%
    \fi
  \fi
    \setlength{\unitlength}{0.0500bp}%
    \ifx\gptboxheight\undefined%
      \newlength{\gptboxheight}%
      \newlength{\gptboxwidth}%
      \newsavebox{\gptboxtext}%
    \fi%
    \setlength{\fboxrule}{0.5pt}%
    \setlength{\fboxsep}{1pt}%
\begin{picture}(4680.00,3276.00)%
    \gplgaddtomacro\gplbacktext{%
      \colorrgb{0.00,0.00,0.00}
      \put(814,811){\makebox(0,0)[r]{\strut{}$0$}}%
      \colorrgb{0.00,0.00,0.00}
      \put(814,1092){\makebox(0,0)[r]{\strut{}$100$}}%
      \colorrgb{0.00,0.00,0.00}
      \put(814,1372){\makebox(0,0)[r]{\strut{}$200$}}%
      \colorrgb{0.00,0.00,0.00}
      \put(814,1653){\makebox(0,0)[r]{\strut{}$300$}}%
      \colorrgb{0.00,0.00,0.00}
      \put(814,1933){\makebox(0,0)[r]{\strut{}$400$}}%
      \colorrgb{0.00,0.00,0.00}
      \put(814,2214){\makebox(0,0)[r]{\strut{}$500$}}%
      \colorrgb{0.00,0.00,0.00}
      \put(814,2494){\makebox(0,0)[r]{\strut{}$600$}}%
      \colorrgb{0.00,0.00,0.00}
      \put(814,2775){\makebox(0,0)[r]{\strut{}$700$}}%
      \colorrgb{0.00,0.00,0.00}
      \put(814,3055){\makebox(0,0)[r]{\strut{}$800$}}%
      \colorrgb{0.00,0.00,0.00}
      \put(993,632){\rotatebox{-45}{\makebox(0,0)[l]{\strut{}$0$}}}%
      \colorrgb{0.00,0.00,0.00}
      \put(1359,632){\rotatebox{-45}{\makebox(0,0)[l]{\strut{}$20$}}}%
      \colorrgb{0.00,0.00,0.00}
      \put(1724,632){\rotatebox{-45}{\makebox(0,0)[l]{\strut{}$40$}}}%
      \colorrgb{0.00,0.00,0.00}
      \put(2090,632){\rotatebox{-45}{\makebox(0,0)[l]{\strut{}$60$}}}%
      \colorrgb{0.00,0.00,0.00}
      \put(2455,632){\rotatebox{-45}{\makebox(0,0)[l]{\strut{}$80$}}}%
      \colorrgb{0.00,0.00,0.00}
      \put(2821,632){\rotatebox{-45}{\makebox(0,0)[l]{\strut{}$100$}}}%
      \colorrgb{0.00,0.00,0.00}
      \put(3186,632){\rotatebox{-45}{\makebox(0,0)[l]{\strut{}$120$}}}%
      \colorrgb{0.00,0.00,0.00}
      \put(3552,632){\rotatebox{-45}{\makebox(0,0)[l]{\strut{}$140$}}}%
      \colorrgb{0.00,0.00,0.00}
      \put(3917,632){\rotatebox{-45}{\makebox(0,0)[l]{\strut{}$160$}}}%
      \colorrgb{0.00,0.00,0.00}
      \put(4283,632){\rotatebox{-45}{\makebox(0,0)[l]{\strut{}$180$}}}%
    }%
    \gplgaddtomacro\gplfronttext{%
      \csname LTb\endcsname
      \put(198,1933){\rotatebox{-270}{\makebox(0,0){\strut{}Count}}}%
      \put(2638,154){\makebox(0,0){\strut{}Cards Created/Edited per Story}}%
      \csname LTb\endcsname
      \put(4119,2775){\makebox(0,0)[r]{\strut{}total=318692}}%
      \put(4119,2606){\makebox(0,0)[r]{\strut{} mean=55.49}}%
      \put(4119,2438){\makebox(0,0)[r]{\strut{} std=56.88}}%
      \put(4119,2270){\makebox(0,0)[r]{\strut{}bin width=29}}%
    }%
    \gplbacktext
    \put(0,0){\includegraphics{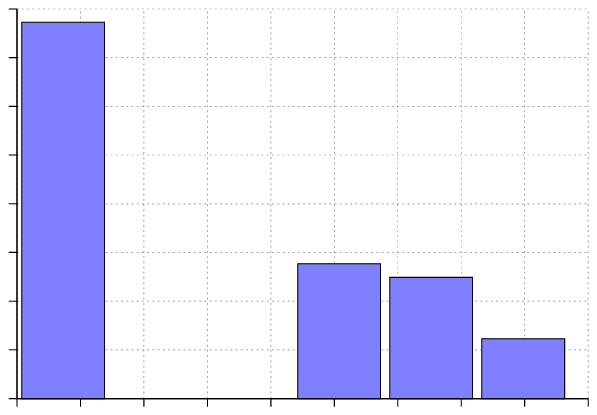}}%
    \gplfronttext
  \end{picture}%
\endgroup
}
\resizebox{\columnwidth}{!}{
\begingroup
  \makeatletter
  \providecommand\color[2][]{%
    \GenericError{(gnuplot) \space\space\space\@spaces}{%
      Package color not loaded in conjunction with
      terminal option `colourtext'%
    }{See the gnuplot documentation for explanation.%
    }{Either use 'blacktext' in gnuplot or load the package
      color.sty in LaTeX.}%
    \renewcommand\color[2][]{}%
  }%
  \providecommand\includegraphics[2][]{%
    \GenericError{(gnuplot) \space\space\space\@spaces}{%
      Package graphicx or graphics not loaded%
    }{See the gnuplot documentation for explanation.%
    }{The gnuplot epslatex terminal needs graphicx.sty or graphics.sty.}%
    \renewcommand\includegraphics[2][]{}%
  }%
  \providecommand\rotatebox[2]{#2}%
  \@ifundefined{ifGPcolor}{%
    \newif\ifGPcolor
    \GPcolortrue
  }{}%
  \@ifundefined{ifGPblacktext}{%
    \newif\ifGPblacktext
    \GPblacktextfalse
  }{}%
  \let\gplgaddtomacro\g@addto@macro
  \gdef\gplbacktext{}%
  \gdef\gplfronttext{}%
  \makeatother
  \ifGPblacktext
    \def\colorrgb#1{}%
    \def\colorgray#1{}%
  \else
    \ifGPcolor
      \def\colorrgb#1{\color[rgb]{#1}}%
      \def\colorgray#1{\color[gray]{#1}}%
      \expandafter\def\csname LTw\endcsname{\color{white}}%
      \expandafter\def\csname LTb\endcsname{\color{black}}%
      \expandafter\def\csname LTa\endcsname{\color{black}}%
      \expandafter\def\csname LT0\endcsname{\color[rgb]{1,0,0}}%
      \expandafter\def\csname LT1\endcsname{\color[rgb]{0,1,0}}%
      \expandafter\def\csname LT2\endcsname{\color[rgb]{0,0,1}}%
      \expandafter\def\csname LT3\endcsname{\color[rgb]{1,0,1}}%
      \expandafter\def\csname LT4\endcsname{\color[rgb]{0,1,1}}%
      \expandafter\def\csname LT5\endcsname{\color[rgb]{1,1,0}}%
      \expandafter\def\csname LT6\endcsname{\color[rgb]{0,0,0}}%
      \expandafter\def\csname LT7\endcsname{\color[rgb]{1,0.3,0}}%
      \expandafter\def\csname LT8\endcsname{\color[rgb]{0.5,0.5,0.5}}%
    \else
      \def\colorrgb#1{\color{black}}%
      \def\colorgray#1{\color[gray]{#1}}%
      \expandafter\def\csname LTw\endcsname{\color{white}}%
      \expandafter\def\csname LTb\endcsname{\color{black}}%
      \expandafter\def\csname LTa\endcsname{\color{black}}%
      \expandafter\def\csname LT0\endcsname{\color{black}}%
      \expandafter\def\csname LT1\endcsname{\color{black}}%
      \expandafter\def\csname LT2\endcsname{\color{black}}%
      \expandafter\def\csname LT3\endcsname{\color{black}}%
      \expandafter\def\csname LT4\endcsname{\color{black}}%
      \expandafter\def\csname LT5\endcsname{\color{black}}%
      \expandafter\def\csname LT6\endcsname{\color{black}}%
      \expandafter\def\csname LT7\endcsname{\color{black}}%
      \expandafter\def\csname LT8\endcsname{\color{black}}%
    \fi
  \fi
    \setlength{\unitlength}{0.0500bp}%
    \ifx\gptboxheight\undefined%
      \newlength{\gptboxheight}%
      \newlength{\gptboxwidth}%
      \newsavebox{\gptboxtext}%
    \fi%
    \setlength{\fboxrule}{0.5pt}%
    \setlength{\fboxsep}{1pt}%
\begin{picture}(4680.00,3276.00)%
    \gplgaddtomacro\gplbacktext{%
      \colorrgb{0.00,0.00,0.00}
      \put(946,811){\makebox(0,0)[r]{\strut{}$32$}}%
      \colorrgb{0.00,0.00,0.00}
      \put(946,1185){\makebox(0,0)[r]{\strut{}$64$}}%
      \colorrgb{0.00,0.00,0.00}
      \put(946,1559){\makebox(0,0)[r]{\strut{}$128$}}%
      \colorrgb{0.00,0.00,0.00}
      \put(946,1933){\makebox(0,0)[r]{\strut{}$256$}}%
      \colorrgb{0.00,0.00,0.00}
      \put(946,2307){\makebox(0,0)[r]{\strut{}$512$}}%
      \colorrgb{0.00,0.00,0.00}
      \put(946,2681){\makebox(0,0)[r]{\strut{}$1024$}}%
      \colorrgb{0.00,0.00,0.00}
      \put(946,3055){\makebox(0,0)[r]{\strut{}$2048$}}%
      \colorrgb{0.00,0.00,0.00}
      \put(1125,632){\rotatebox{-45}{\makebox(0,0)[l]{\strut{}$0$}}}%
      \colorrgb{0.00,0.00,0.00}
      \put(1476,632){\rotatebox{-45}{\makebox(0,0)[l]{\strut{}$20$}}}%
      \colorrgb{0.00,0.00,0.00}
      \put(1827,632){\rotatebox{-45}{\makebox(0,0)[l]{\strut{}$40$}}}%
      \colorrgb{0.00,0.00,0.00}
      \put(2178,632){\rotatebox{-45}{\makebox(0,0)[l]{\strut{}$60$}}}%
      \colorrgb{0.00,0.00,0.00}
      \put(2529,632){\rotatebox{-45}{\makebox(0,0)[l]{\strut{}$80$}}}%
      \colorrgb{0.00,0.00,0.00}
      \put(2879,632){\rotatebox{-45}{\makebox(0,0)[l]{\strut{}$100$}}}%
      \colorrgb{0.00,0.00,0.00}
      \put(3230,632){\rotatebox{-45}{\makebox(0,0)[l]{\strut{}$120$}}}%
      \colorrgb{0.00,0.00,0.00}
      \put(3581,632){\rotatebox{-45}{\makebox(0,0)[l]{\strut{}$140$}}}%
      \colorrgb{0.00,0.00,0.00}
      \put(3932,632){\rotatebox{-45}{\makebox(0,0)[l]{\strut{}$160$}}}%
      \colorrgb{0.00,0.00,0.00}
      \put(4283,632){\rotatebox{-45}{\makebox(0,0)[l]{\strut{}$180$}}}%
    }%
    \gplgaddtomacro\gplfronttext{%
      \csname LTb\endcsname
      \put(198,1933){\rotatebox{-270}{\makebox(0,0){\strut{}Count}}}%
      \put(2704,154){\makebox(0,0){\strut{}Total Played Cards per Story}}%
      \csname LTb\endcsname
      \put(4125,2775){\makebox(0,0)[r]{\strut{}total=232596}}%
      \put(4125,2606){\makebox(0,0)[r]{\strut{} mean=40.50}}%
      \put(4125,2438){\makebox(0,0)[r]{\strut{} std=70.94}}%
      \put(4125,2270){\makebox(0,0)[r]{\strut{}bin width=36}}%
    }%
    \gplbacktext
    \put(0,0){\includegraphics{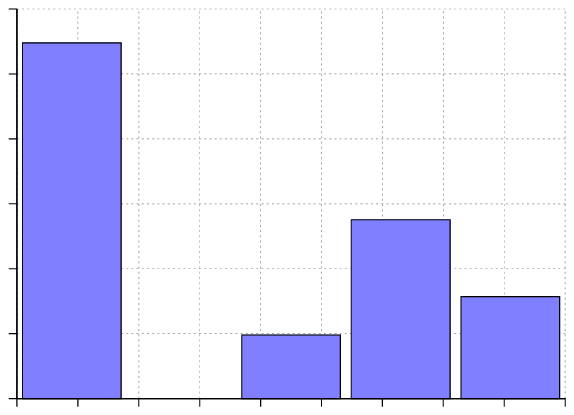}}%
    \gplfronttext
  \end{picture}%
\endgroup
}
\resizebox{\columnwidth}{!}{
\begingroup
  \makeatletter
  \providecommand\color[2][]{%
    \GenericError{(gnuplot) \space\space\space\@spaces}{%
      Package color not loaded in conjunction with
      terminal option `colourtext'%
    }{See the gnuplot documentation for explanation.%
    }{Either use 'blacktext' in gnuplot or load the package
      color.sty in LaTeX.}%
    \renewcommand\color[2][]{}%
  }%
  \providecommand\includegraphics[2][]{%
    \GenericError{(gnuplot) \space\space\space\@spaces}{%
      Package graphicx or graphics not loaded%
    }{See the gnuplot documentation for explanation.%
    }{The gnuplot epslatex terminal needs graphicx.sty or graphics.sty.}%
    \renewcommand\includegraphics[2][]{}%
  }%
  \providecommand\rotatebox[2]{#2}%
  \@ifundefined{ifGPcolor}{%
    \newif\ifGPcolor
    \GPcolortrue
  }{}%
  \@ifundefined{ifGPblacktext}{%
    \newif\ifGPblacktext
    \GPblacktextfalse
  }{}%
  \let\gplgaddtomacro\g@addto@macro
  \gdef\gplbacktext{}%
  \gdef\gplfronttext{}%
  \makeatother
  \ifGPblacktext
    \def\colorrgb#1{}%
    \def\colorgray#1{}%
  \else
    \ifGPcolor
      \def\colorrgb#1{\color[rgb]{#1}}%
      \def\colorgray#1{\color[gray]{#1}}%
      \expandafter\def\csname LTw\endcsname{\color{white}}%
      \expandafter\def\csname LTb\endcsname{\color{black}}%
      \expandafter\def\csname LTa\endcsname{\color{black}}%
      \expandafter\def\csname LT0\endcsname{\color[rgb]{1,0,0}}%
      \expandafter\def\csname LT1\endcsname{\color[rgb]{0,1,0}}%
      \expandafter\def\csname LT2\endcsname{\color[rgb]{0,0,1}}%
      \expandafter\def\csname LT3\endcsname{\color[rgb]{1,0,1}}%
      \expandafter\def\csname LT4\endcsname{\color[rgb]{0,1,1}}%
      \expandafter\def\csname LT5\endcsname{\color[rgb]{1,1,0}}%
      \expandafter\def\csname LT6\endcsname{\color[rgb]{0,0,0}}%
      \expandafter\def\csname LT7\endcsname{\color[rgb]{1,0.3,0}}%
      \expandafter\def\csname LT8\endcsname{\color[rgb]{0.5,0.5,0.5}}%
    \else
      \def\colorrgb#1{\color{black}}%
      \def\colorgray#1{\color[gray]{#1}}%
      \expandafter\def\csname LTw\endcsname{\color{white}}%
      \expandafter\def\csname LTb\endcsname{\color{black}}%
      \expandafter\def\csname LTa\endcsname{\color{black}}%
      \expandafter\def\csname LT0\endcsname{\color{black}}%
      \expandafter\def\csname LT1\endcsname{\color{black}}%
      \expandafter\def\csname LT2\endcsname{\color{black}}%
      \expandafter\def\csname LT3\endcsname{\color{black}}%
      \expandafter\def\csname LT4\endcsname{\color{black}}%
      \expandafter\def\csname LT5\endcsname{\color{black}}%
      \expandafter\def\csname LT6\endcsname{\color{black}}%
      \expandafter\def\csname LT7\endcsname{\color{black}}%
      \expandafter\def\csname LT8\endcsname{\color{black}}%
    \fi
  \fi
    \setlength{\unitlength}{0.0500bp}%
    \ifx\gptboxheight\undefined%
      \newlength{\gptboxheight}%
      \newlength{\gptboxwidth}%
      \newsavebox{\gptboxtext}%
    \fi%
    \setlength{\fboxrule}{0.5pt}%
    \setlength{\fboxsep}{1pt}%
\begin{picture}(4680.00,3276.00)%
    \gplgaddtomacro\gplbacktext{%
      \colorrgb{0.00,0.00,0.00}
      \put(946,811){\makebox(0,0)[r]{\strut{}$8$}}%
      \colorrgb{0.00,0.00,0.00}
      \put(946,1092){\makebox(0,0)[r]{\strut{}$16$}}%
      \colorrgb{0.00,0.00,0.00}
      \put(946,1372){\makebox(0,0)[r]{\strut{}$32$}}%
      \colorrgb{0.00,0.00,0.00}
      \put(946,1652){\makebox(0,0)[r]{\strut{}$64$}}%
      \colorrgb{0.00,0.00,0.00}
      \put(946,1933){\makebox(0,0)[r]{\strut{}$128$}}%
      \colorrgb{0.00,0.00,0.00}
      \put(946,2214){\makebox(0,0)[r]{\strut{}$256$}}%
      \colorrgb{0.00,0.00,0.00}
      \put(946,2494){\makebox(0,0)[r]{\strut{}$512$}}%
      \colorrgb{0.00,0.00,0.00}
      \put(946,2775){\makebox(0,0)[r]{\strut{}$1024$}}%
      \colorrgb{0.00,0.00,0.00}
      \put(946,3055){\makebox(0,0)[r]{\strut{}$2048$}}%
      \colorrgb{0.00,0.00,0.00}
      \put(1125,632){\rotatebox{-45}{\makebox(0,0)[l]{\strut{}$0$}}}%
      \colorrgb{0.00,0.00,0.00}
      \put(1476,632){\rotatebox{-45}{\makebox(0,0)[l]{\strut{}$20$}}}%
      \colorrgb{0.00,0.00,0.00}
      \put(1827,632){\rotatebox{-45}{\makebox(0,0)[l]{\strut{}$40$}}}%
      \colorrgb{0.00,0.00,0.00}
      \put(2178,632){\rotatebox{-45}{\makebox(0,0)[l]{\strut{}$60$}}}%
      \colorrgb{0.00,0.00,0.00}
      \put(2529,632){\rotatebox{-45}{\makebox(0,0)[l]{\strut{}$80$}}}%
      \colorrgb{0.00,0.00,0.00}
      \put(2879,632){\rotatebox{-45}{\makebox(0,0)[l]{\strut{}$100$}}}%
      \colorrgb{0.00,0.00,0.00}
      \put(3230,632){\rotatebox{-45}{\makebox(0,0)[l]{\strut{}$120$}}}%
      \colorrgb{0.00,0.00,0.00}
      \put(3581,632){\rotatebox{-45}{\makebox(0,0)[l]{\strut{}$140$}}}%
      \colorrgb{0.00,0.00,0.00}
      \put(3932,632){\rotatebox{-45}{\makebox(0,0)[l]{\strut{}$160$}}}%
      \colorrgb{0.00,0.00,0.00}
      \put(4283,632){\rotatebox{-45}{\makebox(0,0)[l]{\strut{}$180$}}}%
    }%
    \gplgaddtomacro\gplfronttext{%
      \csname LTb\endcsname
      \put(198,1933){\rotatebox{-270}{\makebox(0,0){\strut{}Count}}}%
      \put(2704,154){\makebox(0,0){\strut{}Played Cards Created/Edited per Story}}%
      \csname LTb\endcsname
      \put(4125,2775){\makebox(0,0)[r]{\strut{}total=204698}}%
      \put(4125,2606){\makebox(0,0)[r]{\strut{} mean=35.64}}%
      \put(4125,2438){\makebox(0,0)[r]{\strut{} std=67.65}}%
      \put(4125,2270){\makebox(0,0)[r]{\strut{}bin width=34}}%
    }%
    \gplbacktext
    \put(0,0){\includegraphics{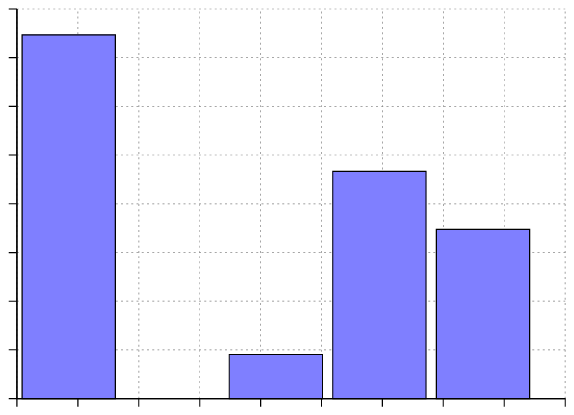}}%
    \gplfronttext
  \end{picture}%
\endgroup
}
\resizebox{\columnwidth}{!}{
\begingroup
  \makeatletter
  \providecommand\color[2][]{%
    \GenericError{(gnuplot) \space\space\space\@spaces}{%
      Package color not loaded in conjunction with
      terminal option `colourtext'%
    }{See the gnuplot documentation for explanation.%
    }{Either use 'blacktext' in gnuplot or load the package
      color.sty in LaTeX.}%
    \renewcommand\color[2][]{}%
  }%
  \providecommand\includegraphics[2][]{%
    \GenericError{(gnuplot) \space\space\space\@spaces}{%
      Package graphicx or graphics not loaded%
    }{See the gnuplot documentation for explanation.%
    }{The gnuplot epslatex terminal needs graphicx.sty or graphics.sty.}%
    \renewcommand\includegraphics[2][]{}%
  }%
  \providecommand\rotatebox[2]{#2}%
  \@ifundefined{ifGPcolor}{%
    \newif\ifGPcolor
    \GPcolortrue
  }{}%
  \@ifundefined{ifGPblacktext}{%
    \newif\ifGPblacktext
    \GPblacktextfalse
  }{}%
  \let\gplgaddtomacro\g@addto@macro
  \gdef\gplbacktext{}%
  \gdef\gplfronttext{}%
  \makeatother
  \ifGPblacktext
    \def\colorrgb#1{}%
    \def\colorgray#1{}%
  \else
    \ifGPcolor
      \def\colorrgb#1{\color[rgb]{#1}}%
      \def\colorgray#1{\color[gray]{#1}}%
      \expandafter\def\csname LTw\endcsname{\color{white}}%
      \expandafter\def\csname LTb\endcsname{\color{black}}%
      \expandafter\def\csname LTa\endcsname{\color{black}}%
      \expandafter\def\csname LT0\endcsname{\color[rgb]{1,0,0}}%
      \expandafter\def\csname LT1\endcsname{\color[rgb]{0,1,0}}%
      \expandafter\def\csname LT2\endcsname{\color[rgb]{0,0,1}}%
      \expandafter\def\csname LT3\endcsname{\color[rgb]{1,0,1}}%
      \expandafter\def\csname LT4\endcsname{\color[rgb]{0,1,1}}%
      \expandafter\def\csname LT5\endcsname{\color[rgb]{1,1,0}}%
      \expandafter\def\csname LT6\endcsname{\color[rgb]{0,0,0}}%
      \expandafter\def\csname LT7\endcsname{\color[rgb]{1,0.3,0}}%
      \expandafter\def\csname LT8\endcsname{\color[rgb]{0.5,0.5,0.5}}%
    \else
      \def\colorrgb#1{\color{black}}%
      \def\colorgray#1{\color[gray]{#1}}%
      \expandafter\def\csname LTw\endcsname{\color{white}}%
      \expandafter\def\csname LTb\endcsname{\color{black}}%
      \expandafter\def\csname LTa\endcsname{\color{black}}%
      \expandafter\def\csname LT0\endcsname{\color{black}}%
      \expandafter\def\csname LT1\endcsname{\color{black}}%
      \expandafter\def\csname LT2\endcsname{\color{black}}%
      \expandafter\def\csname LT3\endcsname{\color{black}}%
      \expandafter\def\csname LT4\endcsname{\color{black}}%
      \expandafter\def\csname LT5\endcsname{\color{black}}%
      \expandafter\def\csname LT6\endcsname{\color{black}}%
      \expandafter\def\csname LT7\endcsname{\color{black}}%
      \expandafter\def\csname LT8\endcsname{\color{black}}%
    \fi
  \fi
    \setlength{\unitlength}{0.0500bp}%
    \ifx\gptboxheight\undefined%
      \newlength{\gptboxheight}%
      \newlength{\gptboxwidth}%
      \newsavebox{\gptboxtext}%
    \fi%
    \setlength{\fboxrule}{0.5pt}%
    \setlength{\fboxsep}{1pt}%
\begin{picture}(4680.00,3276.00)%
    \gplgaddtomacro\gplbacktext{%
      \colorrgb{0.00,0.00,0.00}
      \put(1078,717){\makebox(0,0)[r]{\strut{}$1$}}%
      \colorrgb{0.00,0.00,0.00}
      \put(1078,992){\makebox(0,0)[r]{\strut{}$4$}}%
      \colorrgb{0.00,0.00,0.00}
      \put(1078,1267){\makebox(0,0)[r]{\strut{}$16$}}%
      \colorrgb{0.00,0.00,0.00}
      \put(1078,1542){\makebox(0,0)[r]{\strut{}$64$}}%
      \colorrgb{0.00,0.00,0.00}
      \put(1078,1817){\makebox(0,0)[r]{\strut{}$256$}}%
      \colorrgb{0.00,0.00,0.00}
      \put(1078,2092){\makebox(0,0)[r]{\strut{}$1024$}}%
      \colorrgb{0.00,0.00,0.00}
      \put(1078,2367){\makebox(0,0)[r]{\strut{}$4096$}}%
      \colorrgb{0.00,0.00,0.00}
      \put(1078,2642){\makebox(0,0)[r]{\strut{}$16384$}}%
      \colorrgb{0.00,0.00,0.00}
      \put(1078,2917){\makebox(0,0)[r]{\strut{}$65536$}}%
      \colorrgb{0.00,0.00,0.00}
      \put(1257,538){\rotatebox{-45}{\makebox(0,0)[l]{\strut{}$0$}}}%
      \colorrgb{0.00,0.00,0.00}
      \put(1862,538){\rotatebox{-45}{\makebox(0,0)[l]{\strut{}$5$}}}%
      \colorrgb{0.00,0.00,0.00}
      \put(2467,538){\rotatebox{-45}{\makebox(0,0)[l]{\strut{}$10$}}}%
      \colorrgb{0.00,0.00,0.00}
      \put(3073,538){\rotatebox{-45}{\makebox(0,0)[l]{\strut{}$15$}}}%
      \colorrgb{0.00,0.00,0.00}
      \put(3678,538){\rotatebox{-45}{\makebox(0,0)[l]{\strut{}$20$}}}%
      \colorrgb{0.00,0.00,0.00}
      \put(4283,538){\rotatebox{-45}{\makebox(0,0)[l]{\strut{}$25$}}}%
    }%
    \gplgaddtomacro\gplfronttext{%
      \csname LTb\endcsname
      \put(198,1886){\rotatebox{-270}{\makebox(0,0){\strut{}Count}}}%
      \put(2770,154){\makebox(0,0){\strut{}Challenge Cards per Entry}}%
      \csname LTb\endcsname
      \put(4132,2763){\makebox(0,0)[r]{\strut{}total=61223}}%
      \put(4132,2587){\makebox(0,0)[r]{\strut{} mean=0.47}}%
      \put(4132,2412){\makebox(0,0)[r]{\strut{} std=1.05}}%
      \put(4132,2237){\makebox(0,0)[r]{\strut{}bin width=1}}%
    }%
    \gplbacktext
    \put(0,0){\includegraphics{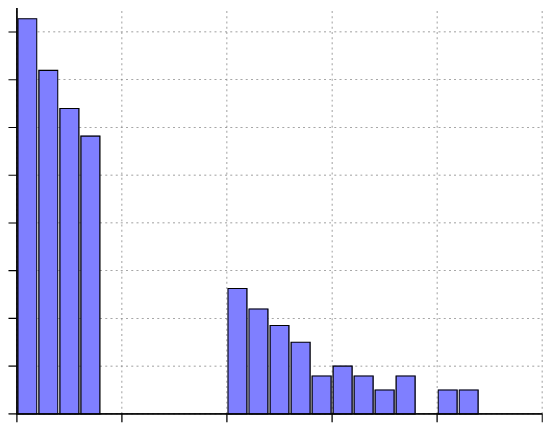}}%
    \gplfronttext
  \end{picture}%
\endgroup
}
\resizebox{\columnwidth}{!}{
\begingroup
  \makeatletter
  \providecommand\color[2][]{%
    \GenericError{(gnuplot) \space\space\space\@spaces}{%
      Package color not loaded in conjunction with
      terminal option `colourtext'%
    }{See the gnuplot documentation for explanation.%
    }{Either use 'blacktext' in gnuplot or load the package
      color.sty in LaTeX.}%
    \renewcommand\color[2][]{}%
  }%
  \providecommand\includegraphics[2][]{%
    \GenericError{(gnuplot) \space\space\space\@spaces}{%
      Package graphicx or graphics not loaded%
    }{See the gnuplot documentation for explanation.%
    }{The gnuplot epslatex terminal needs graphicx.sty or graphics.sty.}%
    \renewcommand\includegraphics[2][]{}%
  }%
  \providecommand\rotatebox[2]{#2}%
  \@ifundefined{ifGPcolor}{%
    \newif\ifGPcolor
    \GPcolortrue
  }{}%
  \@ifundefined{ifGPblacktext}{%
    \newif\ifGPblacktext
    \GPblacktextfalse
  }{}%
  \let\gplgaddtomacro\g@addto@macro
  \gdef\gplbacktext{}%
  \gdef\gplfronttext{}%
  \makeatother
  \ifGPblacktext
    \def\colorrgb#1{}%
    \def\colorgray#1{}%
  \else
    \ifGPcolor
      \def\colorrgb#1{\color[rgb]{#1}}%
      \def\colorgray#1{\color[gray]{#1}}%
      \expandafter\def\csname LTw\endcsname{\color{white}}%
      \expandafter\def\csname LTb\endcsname{\color{black}}%
      \expandafter\def\csname LTa\endcsname{\color{black}}%
      \expandafter\def\csname LT0\endcsname{\color[rgb]{1,0,0}}%
      \expandafter\def\csname LT1\endcsname{\color[rgb]{0,1,0}}%
      \expandafter\def\csname LT2\endcsname{\color[rgb]{0,0,1}}%
      \expandafter\def\csname LT3\endcsname{\color[rgb]{1,0,1}}%
      \expandafter\def\csname LT4\endcsname{\color[rgb]{0,1,1}}%
      \expandafter\def\csname LT5\endcsname{\color[rgb]{1,1,0}}%
      \expandafter\def\csname LT6\endcsname{\color[rgb]{0,0,0}}%
      \expandafter\def\csname LT7\endcsname{\color[rgb]{1,0.3,0}}%
      \expandafter\def\csname LT8\endcsname{\color[rgb]{0.5,0.5,0.5}}%
    \else
      \def\colorrgb#1{\color{black}}%
      \def\colorgray#1{\color[gray]{#1}}%
      \expandafter\def\csname LTw\endcsname{\color{white}}%
      \expandafter\def\csname LTb\endcsname{\color{black}}%
      \expandafter\def\csname LTa\endcsname{\color{black}}%
      \expandafter\def\csname LT0\endcsname{\color{black}}%
      \expandafter\def\csname LT1\endcsname{\color{black}}%
      \expandafter\def\csname LT2\endcsname{\color{black}}%
      \expandafter\def\csname LT3\endcsname{\color{black}}%
      \expandafter\def\csname LT4\endcsname{\color{black}}%
      \expandafter\def\csname LT5\endcsname{\color{black}}%
      \expandafter\def\csname LT6\endcsname{\color{black}}%
      \expandafter\def\csname LT7\endcsname{\color{black}}%
      \expandafter\def\csname LT8\endcsname{\color{black}}%
    \fi
  \fi
    \setlength{\unitlength}{0.0500bp}%
    \ifx\gptboxheight\undefined%
      \newlength{\gptboxheight}%
      \newlength{\gptboxwidth}%
      \newsavebox{\gptboxtext}%
    \fi%
    \setlength{\fboxrule}{0.5pt}%
    \setlength{\fboxsep}{1pt}%
\begin{picture}(4680.00,3276.00)%
    \gplgaddtomacro\gplbacktext{%
      \colorrgb{0.00,0.00,0.00}
      \put(1210,811){\makebox(0,0)[r]{\strut{}$0$}}%
      \colorrgb{0.00,0.00,0.00}
      \put(1210,1035){\makebox(0,0)[r]{\strut{}$20000$}}%
      \colorrgb{0.00,0.00,0.00}
      \put(1210,1260){\makebox(0,0)[r]{\strut{}$40000$}}%
      \colorrgb{0.00,0.00,0.00}
      \put(1210,1484){\makebox(0,0)[r]{\strut{}$60000$}}%
      \colorrgb{0.00,0.00,0.00}
      \put(1210,1709){\makebox(0,0)[r]{\strut{}$80000$}}%
      \colorrgb{0.00,0.00,0.00}
      \put(1210,1933){\makebox(0,0)[r]{\strut{}$100000$}}%
      \colorrgb{0.00,0.00,0.00}
      \put(1210,2157){\makebox(0,0)[r]{\strut{}$120000$}}%
      \colorrgb{0.00,0.00,0.00}
      \put(1210,2382){\makebox(0,0)[r]{\strut{}$140000$}}%
      \colorrgb{0.00,0.00,0.00}
      \put(1210,2606){\makebox(0,0)[r]{\strut{}$160000$}}%
      \colorrgb{0.00,0.00,0.00}
      \put(1210,2831){\makebox(0,0)[r]{\strut{}$180000$}}%
      \colorrgb{0.00,0.00,0.00}
      \put(1210,3055){\makebox(0,0)[r]{\strut{}$200000$}}%
      \colorrgb{0.00,0.00,0.00}
      \put(1389,632){\rotatebox{-45}{\makebox(0,0)[l]{\strut{}$0$}}}%
      \colorrgb{0.00,0.00,0.00}
      \put(1871,632){\rotatebox{-45}{\makebox(0,0)[l]{\strut{}$0.5$}}}%
      \colorrgb{0.00,0.00,0.00}
      \put(2354,632){\rotatebox{-45}{\makebox(0,0)[l]{\strut{}$1$}}}%
      \colorrgb{0.00,0.00,0.00}
      \put(2836,632){\rotatebox{-45}{\makebox(0,0)[l]{\strut{}$1.5$}}}%
      \colorrgb{0.00,0.00,0.00}
      \put(3318,632){\rotatebox{-45}{\makebox(0,0)[l]{\strut{}$2$}}}%
      \colorrgb{0.00,0.00,0.00}
      \put(3801,632){\rotatebox{-45}{\makebox(0,0)[l]{\strut{}$2.5$}}}%
      \colorrgb{0.00,0.00,0.00}
      \put(4283,632){\rotatebox{-45}{\makebox(0,0)[l]{\strut{}$3$}}}%
    }%
    \gplgaddtomacro\gplfronttext{%
      \csname LTb\endcsname
      \put(198,1933){\rotatebox{-270}{\makebox(0,0){\strut{}Count}}}%
      \put(2836,154){\makebox(0,0){\strut{}Played Cards per Entry}}%
      \csname LTb\endcsname
      \put(4138,2775){\makebox(0,0)[r]{\strut{}total=149014}}%
      \put(4138,2606){\makebox(0,0)[r]{\strut{} mean=0.47}}%
      \put(4138,2438){\makebox(0,0)[r]{\strut{} std=0.63}}%
      \put(4138,2270){\makebox(0,0)[r]{\strut{}bin width=1}}%
    }%
    \gplbacktext
    \put(0,0){\includegraphics{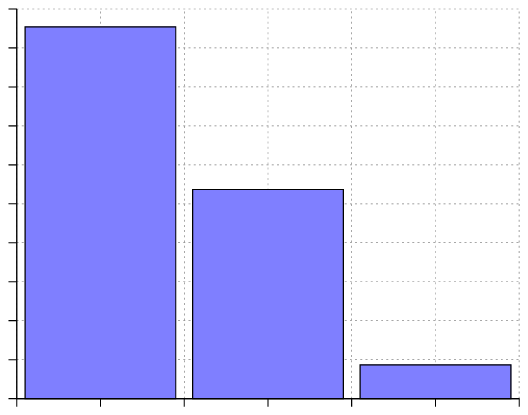}}%
    \gplfronttext
  \end{picture}%
\endgroup
}
\resizebox{\columnwidth}{!}{
\begingroup
  \makeatletter
  \providecommand\color[2][]{%
    \GenericError{(gnuplot) \space\space\space\@spaces}{%
      Package color not loaded in conjunction with
      terminal option `colourtext'%
    }{See the gnuplot documentation for explanation.%
    }{Either use 'blacktext' in gnuplot or load the package
      color.sty in LaTeX.}%
    \renewcommand\color[2][]{}%
  }%
  \providecommand\includegraphics[2][]{%
    \GenericError{(gnuplot) \space\space\space\@spaces}{%
      Package graphicx or graphics not loaded%
    }{See the gnuplot documentation for explanation.%
    }{The gnuplot epslatex terminal needs graphicx.sty or graphics.sty.}%
    \renewcommand\includegraphics[2][]{}%
  }%
  \providecommand\rotatebox[2]{#2}%
  \@ifundefined{ifGPcolor}{%
    \newif\ifGPcolor
    \GPcolortrue
  }{}%
  \@ifundefined{ifGPblacktext}{%
    \newif\ifGPblacktext
    \GPblacktextfalse
  }{}%
  \let\gplgaddtomacro\g@addto@macro
  \gdef\gplbacktext{}%
  \gdef\gplfronttext{}%
  \makeatother
  \ifGPblacktext
    \def\colorrgb#1{}%
    \def\colorgray#1{}%
  \else
    \ifGPcolor
      \def\colorrgb#1{\color[rgb]{#1}}%
      \def\colorgray#1{\color[gray]{#1}}%
      \expandafter\def\csname LTw\endcsname{\color{white}}%
      \expandafter\def\csname LTb\endcsname{\color{black}}%
      \expandafter\def\csname LTa\endcsname{\color{black}}%
      \expandafter\def\csname LT0\endcsname{\color[rgb]{1,0,0}}%
      \expandafter\def\csname LT1\endcsname{\color[rgb]{0,1,0}}%
      \expandafter\def\csname LT2\endcsname{\color[rgb]{0,0,1}}%
      \expandafter\def\csname LT3\endcsname{\color[rgb]{1,0,1}}%
      \expandafter\def\csname LT4\endcsname{\color[rgb]{0,1,1}}%
      \expandafter\def\csname LT5\endcsname{\color[rgb]{1,1,0}}%
      \expandafter\def\csname LT6\endcsname{\color[rgb]{0,0,0}}%
      \expandafter\def\csname LT7\endcsname{\color[rgb]{1,0.3,0}}%
      \expandafter\def\csname LT8\endcsname{\color[rgb]{0.5,0.5,0.5}}%
    \else
      \def\colorrgb#1{\color{black}}%
      \def\colorgray#1{\color[gray]{#1}}%
      \expandafter\def\csname LTw\endcsname{\color{white}}%
      \expandafter\def\csname LTb\endcsname{\color{black}}%
      \expandafter\def\csname LTa\endcsname{\color{black}}%
      \expandafter\def\csname LT0\endcsname{\color{black}}%
      \expandafter\def\csname LT1\endcsname{\color{black}}%
      \expandafter\def\csname LT2\endcsname{\color{black}}%
      \expandafter\def\csname LT3\endcsname{\color{black}}%
      \expandafter\def\csname LT4\endcsname{\color{black}}%
      \expandafter\def\csname LT5\endcsname{\color{black}}%
      \expandafter\def\csname LT6\endcsname{\color{black}}%
      \expandafter\def\csname LT7\endcsname{\color{black}}%
      \expandafter\def\csname LT8\endcsname{\color{black}}%
    \fi
  \fi
    \setlength{\unitlength}{0.0500bp}%
    \ifx\gptboxheight\undefined%
      \newlength{\gptboxheight}%
      \newlength{\gptboxwidth}%
      \newsavebox{\gptboxtext}%
    \fi%
    \setlength{\fboxrule}{0.5pt}%
    \setlength{\fboxsep}{1pt}%
\begin{picture}(4680.00,3276.00)%
    \gplgaddtomacro\gplbacktext{%
      \colorrgb{0.00,0.00,0.00}
      \put(1210,811){\makebox(0,0)[r]{\strut{}$0$}}%
      \colorrgb{0.00,0.00,0.00}
      \put(1210,1185){\makebox(0,0)[r]{\strut{}$50000$}}%
      \colorrgb{0.00,0.00,0.00}
      \put(1210,1559){\makebox(0,0)[r]{\strut{}$100000$}}%
      \colorrgb{0.00,0.00,0.00}
      \put(1210,1933){\makebox(0,0)[r]{\strut{}$150000$}}%
      \colorrgb{0.00,0.00,0.00}
      \put(1210,2307){\makebox(0,0)[r]{\strut{}$200000$}}%
      \colorrgb{0.00,0.00,0.00}
      \put(1210,2681){\makebox(0,0)[r]{\strut{}$250000$}}%
      \colorrgb{0.00,0.00,0.00}
      \put(1210,3055){\makebox(0,0)[r]{\strut{}$300000$}}%
      \colorrgb{0.00,0.00,0.00}
      \put(1389,632){\rotatebox{-45}{\makebox(0,0)[l]{\strut{}$0$}}}%
      \colorrgb{0.00,0.00,0.00}
      \put(1678,632){\rotatebox{-45}{\makebox(0,0)[l]{\strut{}$0.2$}}}%
      \colorrgb{0.00,0.00,0.00}
      \put(1968,632){\rotatebox{-45}{\makebox(0,0)[l]{\strut{}$0.4$}}}%
      \colorrgb{0.00,0.00,0.00}
      \put(2257,632){\rotatebox{-45}{\makebox(0,0)[l]{\strut{}$0.6$}}}%
      \colorrgb{0.00,0.00,0.00}
      \put(2547,632){\rotatebox{-45}{\makebox(0,0)[l]{\strut{}$0.8$}}}%
      \colorrgb{0.00,0.00,0.00}
      \put(2836,632){\rotatebox{-45}{\makebox(0,0)[l]{\strut{}$1$}}}%
      \colorrgb{0.00,0.00,0.00}
      \put(3125,632){\rotatebox{-45}{\makebox(0,0)[l]{\strut{}$1.2$}}}%
      \colorrgb{0.00,0.00,0.00}
      \put(3415,632){\rotatebox{-45}{\makebox(0,0)[l]{\strut{}$1.4$}}}%
      \colorrgb{0.00,0.00,0.00}
      \put(3704,632){\rotatebox{-45}{\makebox(0,0)[l]{\strut{}$1.6$}}}%
      \colorrgb{0.00,0.00,0.00}
      \put(3994,632){\rotatebox{-45}{\makebox(0,0)[l]{\strut{}$1.8$}}}%
      \colorrgb{0.00,0.00,0.00}
      \put(4283,632){\rotatebox{-45}{\makebox(0,0)[l]{\strut{}$2$}}}%
    }%
    \gplgaddtomacro\gplfronttext{%
      \csname LTb\endcsname
      \put(198,1933){\rotatebox{-270}{\makebox(0,0){\strut{}Count}}}%
      \put(2836,154){\makebox(0,0){\strut{}Played Wild Cards per Entry}}%
      \csname LTb\endcsname
      \put(4138,2775){\makebox(0,0)[r]{\strut{}total=31465}}%
      \put(4138,2606){\makebox(0,0)[r]{\strut{} mean=0.10}}%
      \put(4138,2438){\makebox(0,0)[r]{\strut{} std=0.30}}%
      \put(4138,2270){\makebox(0,0)[r]{\strut{}bin width=1}}%
    }%
    \gplbacktext
    \put(0,0){\includegraphics{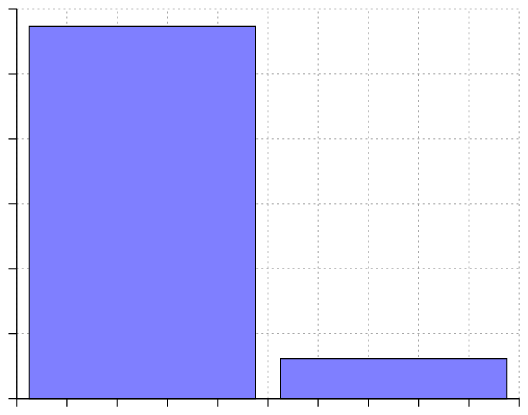}}%
    \gplfronttext
  \end{picture}%
\endgroup
}
\resizebox{\columnwidth}{!}{
\begingroup
  \makeatletter
  \providecommand\color[2][]{%
    \GenericError{(gnuplot) \space\space\space\@spaces}{%
      Package color not loaded in conjunction with
      terminal option `colourtext'%
    }{See the gnuplot documentation for explanation.%
    }{Either use 'blacktext' in gnuplot or load the package
      color.sty in LaTeX.}%
    \renewcommand\color[2][]{}%
  }%
  \providecommand\includegraphics[2][]{%
    \GenericError{(gnuplot) \space\space\space\@spaces}{%
      Package graphicx or graphics not loaded%
    }{See the gnuplot documentation for explanation.%
    }{The gnuplot epslatex terminal needs graphicx.sty or graphics.sty.}%
    \renewcommand\includegraphics[2][]{}%
  }%
  \providecommand\rotatebox[2]{#2}%
  \@ifundefined{ifGPcolor}{%
    \newif\ifGPcolor
    \GPcolortrue
  }{}%
  \@ifundefined{ifGPblacktext}{%
    \newif\ifGPblacktext
    \GPblacktextfalse
  }{}%
  \let\gplgaddtomacro\g@addto@macro
  \gdef\gplbacktext{}%
  \gdef\gplfronttext{}%
  \makeatother
  \ifGPblacktext
    \def\colorrgb#1{}%
    \def\colorgray#1{}%
  \else
    \ifGPcolor
      \def\colorrgb#1{\color[rgb]{#1}}%
      \def\colorgray#1{\color[gray]{#1}}%
      \expandafter\def\csname LTw\endcsname{\color{white}}%
      \expandafter\def\csname LTb\endcsname{\color{black}}%
      \expandafter\def\csname LTa\endcsname{\color{black}}%
      \expandafter\def\csname LT0\endcsname{\color[rgb]{1,0,0}}%
      \expandafter\def\csname LT1\endcsname{\color[rgb]{0,1,0}}%
      \expandafter\def\csname LT2\endcsname{\color[rgb]{0,0,1}}%
      \expandafter\def\csname LT3\endcsname{\color[rgb]{1,0,1}}%
      \expandafter\def\csname LT4\endcsname{\color[rgb]{0,1,1}}%
      \expandafter\def\csname LT5\endcsname{\color[rgb]{1,1,0}}%
      \expandafter\def\csname LT6\endcsname{\color[rgb]{0,0,0}}%
      \expandafter\def\csname LT7\endcsname{\color[rgb]{1,0.3,0}}%
      \expandafter\def\csname LT8\endcsname{\color[rgb]{0.5,0.5,0.5}}%
    \else
      \def\colorrgb#1{\color{black}}%
      \def\colorgray#1{\color[gray]{#1}}%
      \expandafter\def\csname LTw\endcsname{\color{white}}%
      \expandafter\def\csname LTb\endcsname{\color{black}}%
      \expandafter\def\csname LTa\endcsname{\color{black}}%
      \expandafter\def\csname LT0\endcsname{\color{black}}%
      \expandafter\def\csname LT1\endcsname{\color{black}}%
      \expandafter\def\csname LT2\endcsname{\color{black}}%
      \expandafter\def\csname LT3\endcsname{\color{black}}%
      \expandafter\def\csname LT4\endcsname{\color{black}}%
      \expandafter\def\csname LT5\endcsname{\color{black}}%
      \expandafter\def\csname LT6\endcsname{\color{black}}%
      \expandafter\def\csname LT7\endcsname{\color{black}}%
      \expandafter\def\csname LT8\endcsname{\color{black}}%
    \fi
  \fi
    \setlength{\unitlength}{0.0500bp}%
    \ifx\gptboxheight\undefined%
      \newlength{\gptboxheight}%
      \newlength{\gptboxwidth}%
      \newsavebox{\gptboxtext}%
    \fi%
    \setlength{\fboxrule}{0.5pt}%
    \setlength{\fboxsep}{1pt}%
\begin{picture}(4680.00,3276.00)%
    \gplgaddtomacro\gplbacktext{%
      \colorrgb{0.00,0.00,0.00}
      \put(1210,811){\makebox(0,0)[r]{\strut{}$0$}}%
      \colorrgb{0.00,0.00,0.00}
      \put(1210,1260){\makebox(0,0)[r]{\strut{}$50000$}}%
      \colorrgb{0.00,0.00,0.00}
      \put(1210,1709){\makebox(0,0)[r]{\strut{}$100000$}}%
      \colorrgb{0.00,0.00,0.00}
      \put(1210,2157){\makebox(0,0)[r]{\strut{}$150000$}}%
      \colorrgb{0.00,0.00,0.00}
      \put(1210,2606){\makebox(0,0)[r]{\strut{}$200000$}}%
      \colorrgb{0.00,0.00,0.00}
      \put(1210,3055){\makebox(0,0)[r]{\strut{}$250000$}}%
      \colorrgb{0.00,0.00,0.00}
      \put(1389,632){\rotatebox{-45}{\makebox(0,0)[l]{\strut{}$0$}}}%
      \colorrgb{0.00,0.00,0.00}
      \put(1871,632){\rotatebox{-45}{\makebox(0,0)[l]{\strut{}$0.5$}}}%
      \colorrgb{0.00,0.00,0.00}
      \put(2354,632){\rotatebox{-45}{\makebox(0,0)[l]{\strut{}$1$}}}%
      \colorrgb{0.00,0.00,0.00}
      \put(2836,632){\rotatebox{-45}{\makebox(0,0)[l]{\strut{}$1.5$}}}%
      \colorrgb{0.00,0.00,0.00}
      \put(3318,632){\rotatebox{-45}{\makebox(0,0)[l]{\strut{}$2$}}}%
      \colorrgb{0.00,0.00,0.00}
      \put(3801,632){\rotatebox{-45}{\makebox(0,0)[l]{\strut{}$2.5$}}}%
      \colorrgb{0.00,0.00,0.00}
      \put(4283,632){\rotatebox{-45}{\makebox(0,0)[l]{\strut{}$3$}}}%
    }%
    \gplgaddtomacro\gplfronttext{%
      \csname LTb\endcsname
      \put(198,1933){\rotatebox{-270}{\makebox(0,0){\strut{}Count}}}%
      \put(2836,154){\makebox(0,0){\strut{}Played Regular Cards per Entry}}%
      \csname LTb\endcsname
      \put(4138,2775){\makebox(0,0)[r]{\strut{}total=117549}}%
      \put(4138,2606){\makebox(0,0)[r]{\strut{} mean=0.37}}%
      \put(4138,2438){\makebox(0,0)[r]{\strut{} std=0.58}}%
      \put(4138,2270){\makebox(0,0)[r]{\strut{}bin width=1}}%
    }%
    \gplbacktext
    \put(0,0){\includegraphics{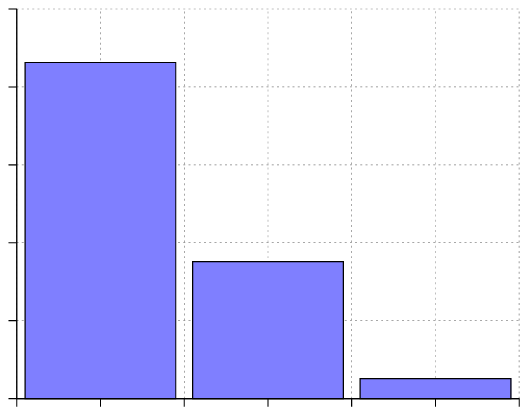}}%
    \gplfronttext
  \end{picture}%
\endgroup
}
\resizebox{\columnwidth}{!}{
\begingroup
  \makeatletter
  \providecommand\color[2][]{%
    \GenericError{(gnuplot) \space\space\space\@spaces}{%
      Package color not loaded in conjunction with
      terminal option `colourtext'%
    }{See the gnuplot documentation for explanation.%
    }{Either use 'blacktext' in gnuplot or load the package
      color.sty in LaTeX.}%
    \renewcommand\color[2][]{}%
  }%
  \providecommand\includegraphics[2][]{%
    \GenericError{(gnuplot) \space\space\space\@spaces}{%
      Package graphicx or graphics not loaded%
    }{See the gnuplot documentation for explanation.%
    }{The gnuplot epslatex terminal needs graphicx.sty or graphics.sty.}%
    \renewcommand\includegraphics[2][]{}%
  }%
  \providecommand\rotatebox[2]{#2}%
  \@ifundefined{ifGPcolor}{%
    \newif\ifGPcolor
    \GPcolortrue
  }{}%
  \@ifundefined{ifGPblacktext}{%
    \newif\ifGPblacktext
    \GPblacktextfalse
  }{}%
  \let\gplgaddtomacro\g@addto@macro
  \gdef\gplbacktext{}%
  \gdef\gplfronttext{}%
  \makeatother
  \ifGPblacktext
    \def\colorrgb#1{}%
    \def\colorgray#1{}%
  \else
    \ifGPcolor
      \def\colorrgb#1{\color[rgb]{#1}}%
      \def\colorgray#1{\color[gray]{#1}}%
      \expandafter\def\csname LTw\endcsname{\color{white}}%
      \expandafter\def\csname LTb\endcsname{\color{black}}%
      \expandafter\def\csname LTa\endcsname{\color{black}}%
      \expandafter\def\csname LT0\endcsname{\color[rgb]{1,0,0}}%
      \expandafter\def\csname LT1\endcsname{\color[rgb]{0,1,0}}%
      \expandafter\def\csname LT2\endcsname{\color[rgb]{0,0,1}}%
      \expandafter\def\csname LT3\endcsname{\color[rgb]{1,0,1}}%
      \expandafter\def\csname LT4\endcsname{\color[rgb]{0,1,1}}%
      \expandafter\def\csname LT5\endcsname{\color[rgb]{1,1,0}}%
      \expandafter\def\csname LT6\endcsname{\color[rgb]{0,0,0}}%
      \expandafter\def\csname LT7\endcsname{\color[rgb]{1,0.3,0}}%
      \expandafter\def\csname LT8\endcsname{\color[rgb]{0.5,0.5,0.5}}%
    \else
      \def\colorrgb#1{\color{black}}%
      \def\colorgray#1{\color[gray]{#1}}%
      \expandafter\def\csname LTw\endcsname{\color{white}}%
      \expandafter\def\csname LTb\endcsname{\color{black}}%
      \expandafter\def\csname LTa\endcsname{\color{black}}%
      \expandafter\def\csname LT0\endcsname{\color{black}}%
      \expandafter\def\csname LT1\endcsname{\color{black}}%
      \expandafter\def\csname LT2\endcsname{\color{black}}%
      \expandafter\def\csname LT3\endcsname{\color{black}}%
      \expandafter\def\csname LT4\endcsname{\color{black}}%
      \expandafter\def\csname LT5\endcsname{\color{black}}%
      \expandafter\def\csname LT6\endcsname{\color{black}}%
      \expandafter\def\csname LT7\endcsname{\color{black}}%
      \expandafter\def\csname LT8\endcsname{\color{black}}%
    \fi
  \fi
    \setlength{\unitlength}{0.0500bp}%
    \ifx\gptboxheight\undefined%
      \newlength{\gptboxheight}%
      \newlength{\gptboxwidth}%
      \newsavebox{\gptboxtext}%
    \fi%
    \setlength{\fboxrule}{0.5pt}%
    \setlength{\fboxsep}{1pt}%
\begin{picture}(4680.00,3276.00)%
    \gplgaddtomacro\gplbacktext{%
      \colorrgb{0.00,0.00,0.00}
      \put(1078,904){\makebox(0,0)[r]{\strut{}$1$}}%
      \colorrgb{0.00,0.00,0.00}
      \put(1078,1211){\makebox(0,0)[r]{\strut{}$4$}}%
      \colorrgb{0.00,0.00,0.00}
      \put(1078,1519){\makebox(0,0)[r]{\strut{}$16$}}%
      \colorrgb{0.00,0.00,0.00}
      \put(1078,1826){\makebox(0,0)[r]{\strut{}$64$}}%
      \colorrgb{0.00,0.00,0.00}
      \put(1078,2133){\makebox(0,0)[r]{\strut{}$256$}}%
      \colorrgb{0.00,0.00,0.00}
      \put(1078,2440){\makebox(0,0)[r]{\strut{}$1024$}}%
      \colorrgb{0.00,0.00,0.00}
      \put(1078,2748){\makebox(0,0)[r]{\strut{}$4096$}}%
      \colorrgb{0.00,0.00,0.00}
      \put(1078,3055){\makebox(0,0)[r]{\strut{}$16384$}}%
      \colorrgb{0.00,0.00,0.00}
      \put(1257,725){\rotatebox{-45}{\makebox(0,0)[l]{\strut{}$0$}}}%
      \colorrgb{0.00,0.00,0.00}
      \put(1862,725){\rotatebox{-45}{\makebox(0,0)[l]{\strut{}$1000$}}}%
      \colorrgb{0.00,0.00,0.00}
      \put(2467,725){\rotatebox{-45}{\makebox(0,0)[l]{\strut{}$2000$}}}%
      \colorrgb{0.00,0.00,0.00}
      \put(3073,725){\rotatebox{-45}{\makebox(0,0)[l]{\strut{}$3000$}}}%
      \colorrgb{0.00,0.00,0.00}
      \put(3678,725){\rotatebox{-45}{\makebox(0,0)[l]{\strut{}$4000$}}}%
      \colorrgb{0.00,0.00,0.00}
      \put(4283,725){\rotatebox{-45}{\makebox(0,0)[l]{\strut{}$5000$}}}%
    }%
    \gplgaddtomacro\gplfronttext{%
      \csname LTb\endcsname
      \put(198,1979){\rotatebox{-270}{\makebox(0,0){\strut{}Count}}}%
      \put(2770,154){\makebox(0,0){\strut{}Tokens per Character Description}}%
      \csname LTb\endcsname
      \put(4132,2786){\makebox(0,0)[r]{\strut{}total=7155548}}%
      \put(4132,2625){\makebox(0,0)[r]{\strut{} mean=260.56}}%
      \put(4132,2463){\makebox(0,0)[r]{\strut{} std=286.78}}%
      \put(4132,2302){\makebox(0,0)[r]{\strut{}bin width=144}}%
    }%
    \gplbacktext
    \put(0,0){\includegraphics{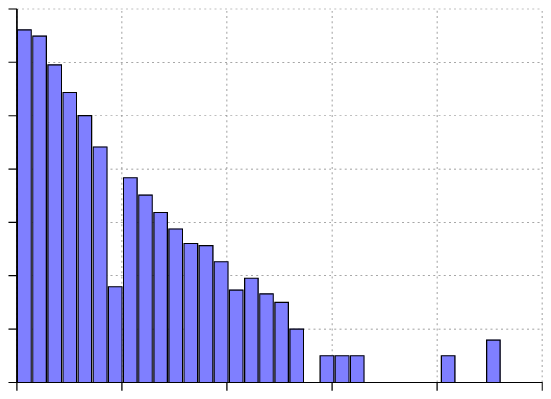}}%
    \gplfronttext
  \end{picture}%
\endgroup
}
\resizebox{\columnwidth}{!}{
\begingroup
  \makeatletter
  \providecommand\color[2][]{%
    \GenericError{(gnuplot) \space\space\space\@spaces}{%
      Package color not loaded in conjunction with
      terminal option `colourtext'%
    }{See the gnuplot documentation for explanation.%
    }{Either use 'blacktext' in gnuplot or load the package
      color.sty in LaTeX.}%
    \renewcommand\color[2][]{}%
  }%
  \providecommand\includegraphics[2][]{%
    \GenericError{(gnuplot) \space\space\space\@spaces}{%
      Package graphicx or graphics not loaded%
    }{See the gnuplot documentation for explanation.%
    }{The gnuplot epslatex terminal needs graphicx.sty or graphics.sty.}%
    \renewcommand\includegraphics[2][]{}%
  }%
  \providecommand\rotatebox[2]{#2}%
  \@ifundefined{ifGPcolor}{%
    \newif\ifGPcolor
    \GPcolortrue
  }{}%
  \@ifundefined{ifGPblacktext}{%
    \newif\ifGPblacktext
    \GPblacktextfalse
  }{}%
  \let\gplgaddtomacro\g@addto@macro
  \gdef\gplbacktext{}%
  \gdef\gplfronttext{}%
  \makeatother
  \ifGPblacktext
    \def\colorrgb#1{}%
    \def\colorgray#1{}%
  \else
    \ifGPcolor
      \def\colorrgb#1{\color[rgb]{#1}}%
      \def\colorgray#1{\color[gray]{#1}}%
      \expandafter\def\csname LTw\endcsname{\color{white}}%
      \expandafter\def\csname LTb\endcsname{\color{black}}%
      \expandafter\def\csname LTa\endcsname{\color{black}}%
      \expandafter\def\csname LT0\endcsname{\color[rgb]{1,0,0}}%
      \expandafter\def\csname LT1\endcsname{\color[rgb]{0,1,0}}%
      \expandafter\def\csname LT2\endcsname{\color[rgb]{0,0,1}}%
      \expandafter\def\csname LT3\endcsname{\color[rgb]{1,0,1}}%
      \expandafter\def\csname LT4\endcsname{\color[rgb]{0,1,1}}%
      \expandafter\def\csname LT5\endcsname{\color[rgb]{1,1,0}}%
      \expandafter\def\csname LT6\endcsname{\color[rgb]{0,0,0}}%
      \expandafter\def\csname LT7\endcsname{\color[rgb]{1,0.3,0}}%
      \expandafter\def\csname LT8\endcsname{\color[rgb]{0.5,0.5,0.5}}%
    \else
      \def\colorrgb#1{\color{black}}%
      \def\colorgray#1{\color[gray]{#1}}%
      \expandafter\def\csname LTw\endcsname{\color{white}}%
      \expandafter\def\csname LTb\endcsname{\color{black}}%
      \expandafter\def\csname LTa\endcsname{\color{black}}%
      \expandafter\def\csname LT0\endcsname{\color{black}}%
      \expandafter\def\csname LT1\endcsname{\color{black}}%
      \expandafter\def\csname LT2\endcsname{\color{black}}%
      \expandafter\def\csname LT3\endcsname{\color{black}}%
      \expandafter\def\csname LT4\endcsname{\color{black}}%
      \expandafter\def\csname LT5\endcsname{\color{black}}%
      \expandafter\def\csname LT6\endcsname{\color{black}}%
      \expandafter\def\csname LT7\endcsname{\color{black}}%
      \expandafter\def\csname LT8\endcsname{\color{black}}%
    \fi
  \fi
    \setlength{\unitlength}{0.0500bp}%
    \ifx\gptboxheight\undefined%
      \newlength{\gptboxheight}%
      \newlength{\gptboxwidth}%
      \newsavebox{\gptboxtext}%
    \fi%
    \setlength{\fboxrule}{0.5pt}%
    \setlength{\fboxsep}{1pt}%
\begin{picture}(4680.00,3276.00)%
    \gplgaddtomacro\gplbacktext{%
      \colorrgb{0.00,0.00,0.00}
      \put(1210,997){\makebox(0,0)[r]{\strut{}$1$}}%
      \colorrgb{0.00,0.00,0.00}
      \put(1210,1226){\makebox(0,0)[r]{\strut{}$4$}}%
      \colorrgb{0.00,0.00,0.00}
      \put(1210,1454){\makebox(0,0)[r]{\strut{}$16$}}%
      \colorrgb{0.00,0.00,0.00}
      \put(1210,1683){\makebox(0,0)[r]{\strut{}$64$}}%
      \colorrgb{0.00,0.00,0.00}
      \put(1210,1912){\makebox(0,0)[r]{\strut{}$256$}}%
      \colorrgb{0.00,0.00,0.00}
      \put(1210,2140){\makebox(0,0)[r]{\strut{}$1024$}}%
      \colorrgb{0.00,0.00,0.00}
      \put(1210,2369){\makebox(0,0)[r]{\strut{}$4096$}}%
      \colorrgb{0.00,0.00,0.00}
      \put(1210,2598){\makebox(0,0)[r]{\strut{}$16384$}}%
      \colorrgb{0.00,0.00,0.00}
      \put(1210,2826){\makebox(0,0)[r]{\strut{}$65536$}}%
      \colorrgb{0.00,0.00,0.00}
      \put(1210,3055){\makebox(0,0)[r]{\strut{}$262144$}}%
      \colorrgb{0.00,0.00,0.00}
      \put(1389,818){\rotatebox{-45}{\makebox(0,0)[l]{\strut{}$0$}}}%
      \colorrgb{0.00,0.00,0.00}
      \put(2113,818){\rotatebox{-45}{\makebox(0,0)[l]{\strut{}$5000$}}}%
      \colorrgb{0.00,0.00,0.00}
      \put(2836,818){\rotatebox{-45}{\makebox(0,0)[l]{\strut{}$10000$}}}%
      \colorrgb{0.00,0.00,0.00}
      \put(3560,818){\rotatebox{-45}{\makebox(0,0)[l]{\strut{}$15000$}}}%
      \colorrgb{0.00,0.00,0.00}
      \put(4283,818){\rotatebox{-45}{\makebox(0,0)[l]{\strut{}$20000$}}}%
    }%
    \gplgaddtomacro\gplfronttext{%
      \csname LTb\endcsname
      \put(198,2026){\rotatebox{-270}{\makebox(0,0){\strut{}Count}}}%
      \put(2836,154){\makebox(0,0){\strut{}Tokens per Scene Entry}}%
      \csname LTb\endcsname
      \put(4138,2798){\makebox(0,0)[r]{\strut{}total=110772426}}%
      \put(4138,2643){\makebox(0,0)[r]{\strut{} mean=247.11}}%
      \put(4138,2489){\makebox(0,0)[r]{\strut{} std=307.13}}%
      \put(4138,2335){\makebox(0,0)[r]{\strut{}bin width=154}}%
    }%
    \gplbacktext
    \put(0,0){\includegraphics{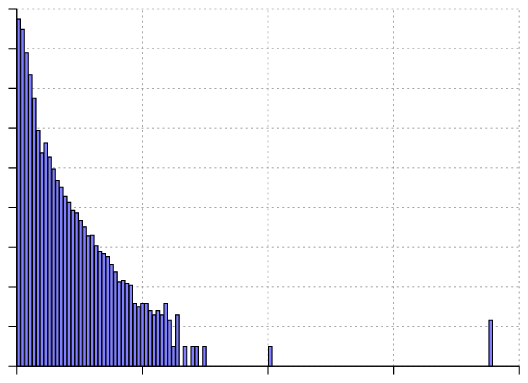}}%
    \gplfronttext
  \end{picture}%
\endgroup
}
\resizebox{\columnwidth}{!}{
\begingroup
  \makeatletter
  \providecommand\color[2][]{%
    \GenericError{(gnuplot) \space\space\space\@spaces}{%
      Package color not loaded in conjunction with
      terminal option `colourtext'%
    }{See the gnuplot documentation for explanation.%
    }{Either use 'blacktext' in gnuplot or load the package
      color.sty in LaTeX.}%
    \renewcommand\color[2][]{}%
  }%
  \providecommand\includegraphics[2][]{%
    \GenericError{(gnuplot) \space\space\space\@spaces}{%
      Package graphicx or graphics not loaded%
    }{See the gnuplot documentation for explanation.%
    }{The gnuplot epslatex terminal needs graphicx.sty or graphics.sty.}%
    \renewcommand\includegraphics[2][]{}%
  }%
  \providecommand\rotatebox[2]{#2}%
  \@ifundefined{ifGPcolor}{%
    \newif\ifGPcolor
    \GPcolortrue
  }{}%
  \@ifundefined{ifGPblacktext}{%
    \newif\ifGPblacktext
    \GPblacktextfalse
  }{}%
  \let\gplgaddtomacro\g@addto@macro
  \gdef\gplbacktext{}%
  \gdef\gplfronttext{}%
  \makeatother
  \ifGPblacktext
    \def\colorrgb#1{}%
    \def\colorgray#1{}%
  \else
    \ifGPcolor
      \def\colorrgb#1{\color[rgb]{#1}}%
      \def\colorgray#1{\color[gray]{#1}}%
      \expandafter\def\csname LTw\endcsname{\color{white}}%
      \expandafter\def\csname LTb\endcsname{\color{black}}%
      \expandafter\def\csname LTa\endcsname{\color{black}}%
      \expandafter\def\csname LT0\endcsname{\color[rgb]{1,0,0}}%
      \expandafter\def\csname LT1\endcsname{\color[rgb]{0,1,0}}%
      \expandafter\def\csname LT2\endcsname{\color[rgb]{0,0,1}}%
      \expandafter\def\csname LT3\endcsname{\color[rgb]{1,0,1}}%
      \expandafter\def\csname LT4\endcsname{\color[rgb]{0,1,1}}%
      \expandafter\def\csname LT5\endcsname{\color[rgb]{1,1,0}}%
      \expandafter\def\csname LT6\endcsname{\color[rgb]{0,0,0}}%
      \expandafter\def\csname LT7\endcsname{\color[rgb]{1,0.3,0}}%
      \expandafter\def\csname LT8\endcsname{\color[rgb]{0.5,0.5,0.5}}%
    \else
      \def\colorrgb#1{\color{black}}%
      \def\colorgray#1{\color[gray]{#1}}%
      \expandafter\def\csname LTw\endcsname{\color{white}}%
      \expandafter\def\csname LTb\endcsname{\color{black}}%
      \expandafter\def\csname LTa\endcsname{\color{black}}%
      \expandafter\def\csname LT0\endcsname{\color{black}}%
      \expandafter\def\csname LT1\endcsname{\color{black}}%
      \expandafter\def\csname LT2\endcsname{\color{black}}%
      \expandafter\def\csname LT3\endcsname{\color{black}}%
      \expandafter\def\csname LT4\endcsname{\color{black}}%
      \expandafter\def\csname LT5\endcsname{\color{black}}%
      \expandafter\def\csname LT6\endcsname{\color{black}}%
      \expandafter\def\csname LT7\endcsname{\color{black}}%
      \expandafter\def\csname LT8\endcsname{\color{black}}%
    \fi
  \fi
    \setlength{\unitlength}{0.0500bp}%
    \ifx\gptboxheight\undefined%
      \newlength{\gptboxheight}%
      \newlength{\gptboxwidth}%
      \newsavebox{\gptboxtext}%
    \fi%
    \setlength{\fboxrule}{0.5pt}%
    \setlength{\fboxsep}{1pt}%
\begin{picture}(4680.00,3276.00)%
    \gplgaddtomacro\gplbacktext{%
      \colorrgb{0.00,0.00,0.00}
      \put(1210,936){\makebox(0,0)[r]{\strut{}$4$}}%
      \colorrgb{0.00,0.00,0.00}
      \put(1210,1185){\makebox(0,0)[r]{\strut{}$16$}}%
      \colorrgb{0.00,0.00,0.00}
      \put(1210,1434){\makebox(0,0)[r]{\strut{}$64$}}%
      \colorrgb{0.00,0.00,0.00}
      \put(1210,1684){\makebox(0,0)[r]{\strut{}$256$}}%
      \colorrgb{0.00,0.00,0.00}
      \put(1210,1933){\makebox(0,0)[r]{\strut{}$1024$}}%
      \colorrgb{0.00,0.00,0.00}
      \put(1210,2182){\makebox(0,0)[r]{\strut{}$4096$}}%
      \colorrgb{0.00,0.00,0.00}
      \put(1210,2432){\makebox(0,0)[r]{\strut{}$16384$}}%
      \colorrgb{0.00,0.00,0.00}
      \put(1210,2681){\makebox(0,0)[r]{\strut{}$65536$}}%
      \colorrgb{0.00,0.00,0.00}
      \put(1210,2930){\makebox(0,0)[r]{\strut{}$262144$}}%
      \colorrgb{0.00,0.00,0.00}
      \put(1389,632){\rotatebox{-45}{\makebox(0,0)[l]{\strut{}$0$}}}%
      \colorrgb{0.00,0.00,0.00}
      \put(1802,632){\rotatebox{-45}{\makebox(0,0)[l]{\strut{}$20$}}}%
      \colorrgb{0.00,0.00,0.00}
      \put(2216,632){\rotatebox{-45}{\makebox(0,0)[l]{\strut{}$40$}}}%
      \colorrgb{0.00,0.00,0.00}
      \put(2629,632){\rotatebox{-45}{\makebox(0,0)[l]{\strut{}$60$}}}%
      \colorrgb{0.00,0.00,0.00}
      \put(3043,632){\rotatebox{-45}{\makebox(0,0)[l]{\strut{}$80$}}}%
      \colorrgb{0.00,0.00,0.00}
      \put(3456,632){\rotatebox{-45}{\makebox(0,0)[l]{\strut{}$100$}}}%
      \colorrgb{0.00,0.00,0.00}
      \put(3870,632){\rotatebox{-45}{\makebox(0,0)[l]{\strut{}$120$}}}%
      \colorrgb{0.00,0.00,0.00}
      \put(4283,632){\rotatebox{-45}{\makebox(0,0)[l]{\strut{}$140$}}}%
    }%
    \gplgaddtomacro\gplfronttext{%
      \csname LTb\endcsname
      \put(198,1933){\rotatebox{-270}{\makebox(0,0){\strut{}Count}}}%
      \put(2836,154){\makebox(0,0){\strut{}Tokens per Played Location Card}}%
      \csname LTb\endcsname
      \put(4138,2775){\makebox(0,0)[r]{\strut{}total=438044}}%
      \put(4138,2606){\makebox(0,0)[r]{\strut{} mean=0.98}}%
      \put(4138,2438){\makebox(0,0)[r]{\strut{} std=6.14}}%
      \put(4138,2270){\makebox(0,0)[r]{\strut{}bin width=4}}%
    }%
    \gplbacktext
    \put(0,0){\includegraphics{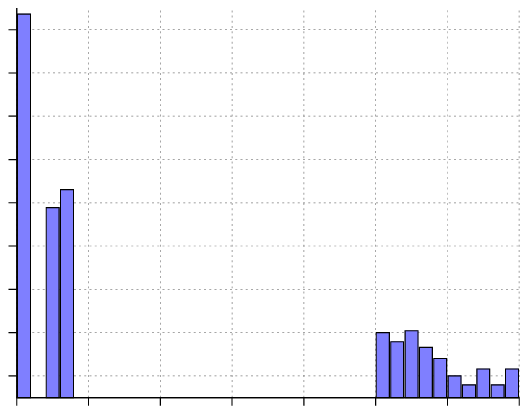}}%
    \gplfronttext
  \end{picture}%
\endgroup
}
\resizebox{\columnwidth}{!}{
\begingroup
  \makeatletter
  \providecommand\color[2][]{%
    \GenericError{(gnuplot) \space\space\space\@spaces}{%
      Package color not loaded in conjunction with
      terminal option `colourtext'%
    }{See the gnuplot documentation for explanation.%
    }{Either use 'blacktext' in gnuplot or load the package
      color.sty in LaTeX.}%
    \renewcommand\color[2][]{}%
  }%
  \providecommand\includegraphics[2][]{%
    \GenericError{(gnuplot) \space\space\space\@spaces}{%
      Package graphicx or graphics not loaded%
    }{See the gnuplot documentation for explanation.%
    }{The gnuplot epslatex terminal needs graphicx.sty or graphics.sty.}%
    \renewcommand\includegraphics[2][]{}%
  }%
  \providecommand\rotatebox[2]{#2}%
  \@ifundefined{ifGPcolor}{%
    \newif\ifGPcolor
    \GPcolortrue
  }{}%
  \@ifundefined{ifGPblacktext}{%
    \newif\ifGPblacktext
    \GPblacktextfalse
  }{}%
  \let\gplgaddtomacro\g@addto@macro
  \gdef\gplbacktext{}%
  \gdef\gplfronttext{}%
  \makeatother
  \ifGPblacktext
    \def\colorrgb#1{}%
    \def\colorgray#1{}%
  \else
    \ifGPcolor
      \def\colorrgb#1{\color[rgb]{#1}}%
      \def\colorgray#1{\color[gray]{#1}}%
      \expandafter\def\csname LTw\endcsname{\color{white}}%
      \expandafter\def\csname LTb\endcsname{\color{black}}%
      \expandafter\def\csname LTa\endcsname{\color{black}}%
      \expandafter\def\csname LT0\endcsname{\color[rgb]{1,0,0}}%
      \expandafter\def\csname LT1\endcsname{\color[rgb]{0,1,0}}%
      \expandafter\def\csname LT2\endcsname{\color[rgb]{0,0,1}}%
      \expandafter\def\csname LT3\endcsname{\color[rgb]{1,0,1}}%
      \expandafter\def\csname LT4\endcsname{\color[rgb]{0,1,1}}%
      \expandafter\def\csname LT5\endcsname{\color[rgb]{1,1,0}}%
      \expandafter\def\csname LT6\endcsname{\color[rgb]{0,0,0}}%
      \expandafter\def\csname LT7\endcsname{\color[rgb]{1,0.3,0}}%
      \expandafter\def\csname LT8\endcsname{\color[rgb]{0.5,0.5,0.5}}%
    \else
      \def\colorrgb#1{\color{black}}%
      \def\colorgray#1{\color[gray]{#1}}%
      \expandafter\def\csname LTw\endcsname{\color{white}}%
      \expandafter\def\csname LTb\endcsname{\color{black}}%
      \expandafter\def\csname LTa\endcsname{\color{black}}%
      \expandafter\def\csname LT0\endcsname{\color{black}}%
      \expandafter\def\csname LT1\endcsname{\color{black}}%
      \expandafter\def\csname LT2\endcsname{\color{black}}%
      \expandafter\def\csname LT3\endcsname{\color{black}}%
      \expandafter\def\csname LT4\endcsname{\color{black}}%
      \expandafter\def\csname LT5\endcsname{\color{black}}%
      \expandafter\def\csname LT6\endcsname{\color{black}}%
      \expandafter\def\csname LT7\endcsname{\color{black}}%
      \expandafter\def\csname LT8\endcsname{\color{black}}%
    \fi
  \fi
    \setlength{\unitlength}{0.0500bp}%
    \ifx\gptboxheight\undefined%
      \newlength{\gptboxheight}%
      \newlength{\gptboxwidth}%
      \newsavebox{\gptboxtext}%
    \fi%
    \setlength{\fboxrule}{0.5pt}%
    \setlength{\fboxsep}{1pt}%
\begin{picture}(4680.00,3276.00)%
    \gplgaddtomacro\gplbacktext{%
      \colorrgb{0.00,0.00,0.00}
      \put(1078,811){\makebox(0,0)[r]{\strut{}$1$}}%
      \colorrgb{0.00,0.00,0.00}
      \put(1078,1092){\makebox(0,0)[r]{\strut{}$4$}}%
      \colorrgb{0.00,0.00,0.00}
      \put(1078,1372){\makebox(0,0)[r]{\strut{}$16$}}%
      \colorrgb{0.00,0.00,0.00}
      \put(1078,1653){\makebox(0,0)[r]{\strut{}$64$}}%
      \colorrgb{0.00,0.00,0.00}
      \put(1078,1933){\makebox(0,0)[r]{\strut{}$256$}}%
      \colorrgb{0.00,0.00,0.00}
      \put(1078,2214){\makebox(0,0)[r]{\strut{}$1024$}}%
      \colorrgb{0.00,0.00,0.00}
      \put(1078,2494){\makebox(0,0)[r]{\strut{}$4096$}}%
      \colorrgb{0.00,0.00,0.00}
      \put(1078,2775){\makebox(0,0)[r]{\strut{}$16384$}}%
      \colorrgb{0.00,0.00,0.00}
      \put(1078,3055){\makebox(0,0)[r]{\strut{}$65536$}}%
      \colorrgb{0.00,0.00,0.00}
      \put(1257,632){\rotatebox{-45}{\makebox(0,0)[l]{\strut{}$0$}}}%
      \colorrgb{0.00,0.00,0.00}
      \put(1929,632){\rotatebox{-45}{\makebox(0,0)[l]{\strut{}$50$}}}%
      \colorrgb{0.00,0.00,0.00}
      \put(2602,632){\rotatebox{-45}{\makebox(0,0)[l]{\strut{}$100$}}}%
      \colorrgb{0.00,0.00,0.00}
      \put(3274,632){\rotatebox{-45}{\makebox(0,0)[l]{\strut{}$150$}}}%
      \colorrgb{0.00,0.00,0.00}
      \put(3947,632){\rotatebox{-45}{\makebox(0,0)[l]{\strut{}$200$}}}%
    }%
    \gplgaddtomacro\gplfronttext{%
      \csname LTb\endcsname
      \put(198,1933){\rotatebox{-270}{\makebox(0,0){\strut{}Count}}}%
      \put(2770,154){\makebox(0,0){\strut{}Tokens per Played Challenge Card}}%
      \csname LTb\endcsname
      \put(4132,2775){\makebox(0,0)[r]{\strut{}total=3837860}}%
      \put(4132,2606){\makebox(0,0)[r]{\strut{} mean=24.84}}%
      \put(4132,2438){\makebox(0,0)[r]{\strut{} std=15.30}}%
      \put(4132,2270){\makebox(0,0)[r]{\strut{}bin width=8}}%
    }%
    \gplbacktext
    \put(0,0){\includegraphics{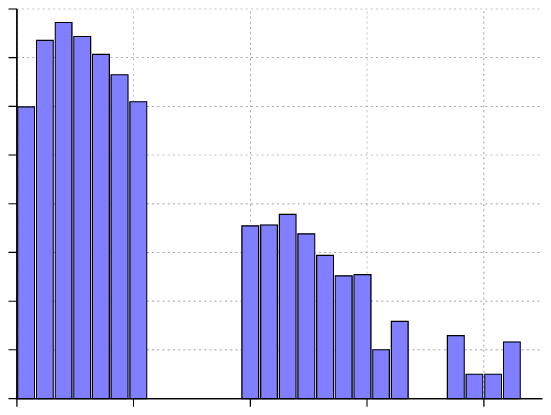}}%
    \gplfronttext
  \end{picture}%
\endgroup
}
\resizebox{\columnwidth}{!}{
\begingroup
  \makeatletter
  \providecommand\color[2][]{%
    \GenericError{(gnuplot) \space\space\space\@spaces}{%
      Package color not loaded in conjunction with
      terminal option `colourtext'%
    }{See the gnuplot documentation for explanation.%
    }{Either use 'blacktext' in gnuplot or load the package
      color.sty in LaTeX.}%
    \renewcommand\color[2][]{}%
  }%
  \providecommand\includegraphics[2][]{%
    \GenericError{(gnuplot) \space\space\space\@spaces}{%
      Package graphicx or graphics not loaded%
    }{See the gnuplot documentation for explanation.%
    }{The gnuplot epslatex terminal needs graphicx.sty or graphics.sty.}%
    \renewcommand\includegraphics[2][]{}%
  }%
  \providecommand\rotatebox[2]{#2}%
  \@ifundefined{ifGPcolor}{%
    \newif\ifGPcolor
    \GPcolortrue
  }{}%
  \@ifundefined{ifGPblacktext}{%
    \newif\ifGPblacktext
    \GPblacktextfalse
  }{}%
  \let\gplgaddtomacro\g@addto@macro
  \gdef\gplbacktext{}%
  \gdef\gplfronttext{}%
  \makeatother
  \ifGPblacktext
    \def\colorrgb#1{}%
    \def\colorgray#1{}%
  \else
    \ifGPcolor
      \def\colorrgb#1{\color[rgb]{#1}}%
      \def\colorgray#1{\color[gray]{#1}}%
      \expandafter\def\csname LTw\endcsname{\color{white}}%
      \expandafter\def\csname LTb\endcsname{\color{black}}%
      \expandafter\def\csname LTa\endcsname{\color{black}}%
      \expandafter\def\csname LT0\endcsname{\color[rgb]{1,0,0}}%
      \expandafter\def\csname LT1\endcsname{\color[rgb]{0,1,0}}%
      \expandafter\def\csname LT2\endcsname{\color[rgb]{0,0,1}}%
      \expandafter\def\csname LT3\endcsname{\color[rgb]{1,0,1}}%
      \expandafter\def\csname LT4\endcsname{\color[rgb]{0,1,1}}%
      \expandafter\def\csname LT5\endcsname{\color[rgb]{1,1,0}}%
      \expandafter\def\csname LT6\endcsname{\color[rgb]{0,0,0}}%
      \expandafter\def\csname LT7\endcsname{\color[rgb]{1,0.3,0}}%
      \expandafter\def\csname LT8\endcsname{\color[rgb]{0.5,0.5,0.5}}%
    \else
      \def\colorrgb#1{\color{black}}%
      \def\colorgray#1{\color[gray]{#1}}%
      \expandafter\def\csname LTw\endcsname{\color{white}}%
      \expandafter\def\csname LTb\endcsname{\color{black}}%
      \expandafter\def\csname LTa\endcsname{\color{black}}%
      \expandafter\def\csname LT0\endcsname{\color{black}}%
      \expandafter\def\csname LT1\endcsname{\color{black}}%
      \expandafter\def\csname LT2\endcsname{\color{black}}%
      \expandafter\def\csname LT3\endcsname{\color{black}}%
      \expandafter\def\csname LT4\endcsname{\color{black}}%
      \expandafter\def\csname LT5\endcsname{\color{black}}%
      \expandafter\def\csname LT6\endcsname{\color{black}}%
      \expandafter\def\csname LT7\endcsname{\color{black}}%
      \expandafter\def\csname LT8\endcsname{\color{black}}%
    \fi
  \fi
    \setlength{\unitlength}{0.0500bp}%
    \ifx\gptboxheight\undefined%
      \newlength{\gptboxheight}%
      \newlength{\gptboxwidth}%
      \newsavebox{\gptboxtext}%
    \fi%
    \setlength{\fboxrule}{0.5pt}%
    \setlength{\fboxsep}{1pt}%
\begin{picture}(4680.00,3276.00)%
    \gplgaddtomacro\gplbacktext{%
      \colorrgb{0.00,0.00,0.00}
      \put(1078,811){\makebox(0,0)[r]{\strut{}$1$}}%
      \colorrgb{0.00,0.00,0.00}
      \put(1078,1092){\makebox(0,0)[r]{\strut{}$4$}}%
      \colorrgb{0.00,0.00,0.00}
      \put(1078,1372){\makebox(0,0)[r]{\strut{}$16$}}%
      \colorrgb{0.00,0.00,0.00}
      \put(1078,1653){\makebox(0,0)[r]{\strut{}$64$}}%
      \colorrgb{0.00,0.00,0.00}
      \put(1078,1933){\makebox(0,0)[r]{\strut{}$256$}}%
      \colorrgb{0.00,0.00,0.00}
      \put(1078,2214){\makebox(0,0)[r]{\strut{}$1024$}}%
      \colorrgb{0.00,0.00,0.00}
      \put(1078,2494){\makebox(0,0)[r]{\strut{}$4096$}}%
      \colorrgb{0.00,0.00,0.00}
      \put(1078,2775){\makebox(0,0)[r]{\strut{}$16384$}}%
      \colorrgb{0.00,0.00,0.00}
      \put(1078,3055){\makebox(0,0)[r]{\strut{}$65536$}}%
      \colorrgb{0.00,0.00,0.00}
      \put(1257,632){\rotatebox{-45}{\makebox(0,0)[l]{\strut{}$0$}}}%
      \colorrgb{0.00,0.00,0.00}
      \put(1929,632){\rotatebox{-45}{\makebox(0,0)[l]{\strut{}$50$}}}%
      \colorrgb{0.00,0.00,0.00}
      \put(2602,632){\rotatebox{-45}{\makebox(0,0)[l]{\strut{}$100$}}}%
      \colorrgb{0.00,0.00,0.00}
      \put(3274,632){\rotatebox{-45}{\makebox(0,0)[l]{\strut{}$150$}}}%
      \colorrgb{0.00,0.00,0.00}
      \put(3947,632){\rotatebox{-45}{\makebox(0,0)[l]{\strut{}$200$}}}%
    }%
    \gplgaddtomacro\gplfronttext{%
      \csname LTb\endcsname
      \put(198,1933){\rotatebox{-270}{\makebox(0,0){\strut{}Count}}}%
      \put(2770,154){\makebox(0,0){\strut{}Tokens per Played Regular Card}}%
      \csname LTb\endcsname
      \put(4132,2775){\makebox(0,0)[r]{\strut{}total=3053152}}%
      \put(4132,2606){\makebox(0,0)[r]{\strut{} mean=25.08}}%
      \put(4132,2438){\makebox(0,0)[r]{\strut{} std=15.78}}%
      \put(4132,2270){\makebox(0,0)[r]{\strut{}bin width=8}}%
    }%
    \gplbacktext
    \put(0,0){\includegraphics{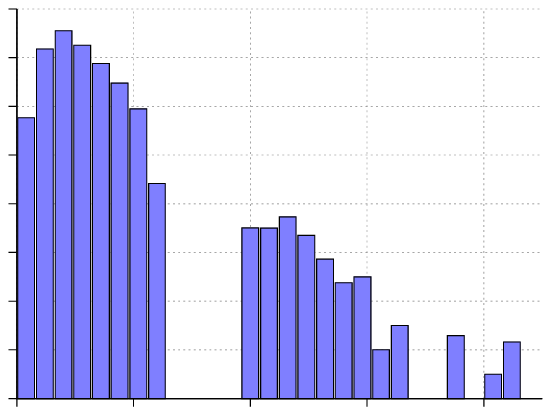}}%
    \gplfronttext
  \end{picture}%
\endgroup
}
\resizebox{\columnwidth}{!}{
\begingroup
  \makeatletter
  \providecommand\color[2][]{%
    \GenericError{(gnuplot) \space\space\space\@spaces}{%
      Package color not loaded in conjunction with
      terminal option `colourtext'%
    }{See the gnuplot documentation for explanation.%
    }{Either use 'blacktext' in gnuplot or load the package
      color.sty in LaTeX.}%
    \renewcommand\color[2][]{}%
  }%
  \providecommand\includegraphics[2][]{%
    \GenericError{(gnuplot) \space\space\space\@spaces}{%
      Package graphicx or graphics not loaded%
    }{See the gnuplot documentation for explanation.%
    }{The gnuplot epslatex terminal needs graphicx.sty or graphics.sty.}%
    \renewcommand\includegraphics[2][]{}%
  }%
  \providecommand\rotatebox[2]{#2}%
  \@ifundefined{ifGPcolor}{%
    \newif\ifGPcolor
    \GPcolortrue
  }{}%
  \@ifundefined{ifGPblacktext}{%
    \newif\ifGPblacktext
    \GPblacktextfalse
  }{}%
  \let\gplgaddtomacro\g@addto@macro
  \gdef\gplbacktext{}%
  \gdef\gplfronttext{}%
  \makeatother
  \ifGPblacktext
    \def\colorrgb#1{}%
    \def\colorgray#1{}%
  \else
    \ifGPcolor
      \def\colorrgb#1{\color[rgb]{#1}}%
      \def\colorgray#1{\color[gray]{#1}}%
      \expandafter\def\csname LTw\endcsname{\color{white}}%
      \expandafter\def\csname LTb\endcsname{\color{black}}%
      \expandafter\def\csname LTa\endcsname{\color{black}}%
      \expandafter\def\csname LT0\endcsname{\color[rgb]{1,0,0}}%
      \expandafter\def\csname LT1\endcsname{\color[rgb]{0,1,0}}%
      \expandafter\def\csname LT2\endcsname{\color[rgb]{0,0,1}}%
      \expandafter\def\csname LT3\endcsname{\color[rgb]{1,0,1}}%
      \expandafter\def\csname LT4\endcsname{\color[rgb]{0,1,1}}%
      \expandafter\def\csname LT5\endcsname{\color[rgb]{1,1,0}}%
      \expandafter\def\csname LT6\endcsname{\color[rgb]{0,0,0}}%
      \expandafter\def\csname LT7\endcsname{\color[rgb]{1,0.3,0}}%
      \expandafter\def\csname LT8\endcsname{\color[rgb]{0.5,0.5,0.5}}%
    \else
      \def\colorrgb#1{\color{black}}%
      \def\colorgray#1{\color[gray]{#1}}%
      \expandafter\def\csname LTw\endcsname{\color{white}}%
      \expandafter\def\csname LTb\endcsname{\color{black}}%
      \expandafter\def\csname LTa\endcsname{\color{black}}%
      \expandafter\def\csname LT0\endcsname{\color{black}}%
      \expandafter\def\csname LT1\endcsname{\color{black}}%
      \expandafter\def\csname LT2\endcsname{\color{black}}%
      \expandafter\def\csname LT3\endcsname{\color{black}}%
      \expandafter\def\csname LT4\endcsname{\color{black}}%
      \expandafter\def\csname LT5\endcsname{\color{black}}%
      \expandafter\def\csname LT6\endcsname{\color{black}}%
      \expandafter\def\csname LT7\endcsname{\color{black}}%
      \expandafter\def\csname LT8\endcsname{\color{black}}%
    \fi
  \fi
    \setlength{\unitlength}{0.0500bp}%
    \ifx\gptboxheight\undefined%
      \newlength{\gptboxheight}%
      \newlength{\gptboxwidth}%
      \newsavebox{\gptboxtext}%
    \fi%
    \setlength{\fboxrule}{0.5pt}%
    \setlength{\fboxsep}{1pt}%
\begin{picture}(4680.00,3276.00)%
    \gplgaddtomacro\gplbacktext{%
      \colorrgb{0.00,0.00,0.00}
      \put(946,811){\makebox(0,0)[r]{\strut{}$1$}}%
      \colorrgb{0.00,0.00,0.00}
      \put(946,1156){\makebox(0,0)[r]{\strut{}$4$}}%
      \colorrgb{0.00,0.00,0.00}
      \put(946,1501){\makebox(0,0)[r]{\strut{}$16$}}%
      \colorrgb{0.00,0.00,0.00}
      \put(946,1847){\makebox(0,0)[r]{\strut{}$64$}}%
      \colorrgb{0.00,0.00,0.00}
      \put(946,2192){\makebox(0,0)[r]{\strut{}$256$}}%
      \colorrgb{0.00,0.00,0.00}
      \put(946,2537){\makebox(0,0)[r]{\strut{}$1024$}}%
      \colorrgb{0.00,0.00,0.00}
      \put(946,2882){\makebox(0,0)[r]{\strut{}$4096$}}%
      \colorrgb{0.00,0.00,0.00}
      \put(1125,632){\rotatebox{-45}{\makebox(0,0)[l]{\strut{}$0$}}}%
      \colorrgb{0.00,0.00,0.00}
      \put(1520,632){\rotatebox{-45}{\makebox(0,0)[l]{\strut{}$20$}}}%
      \colorrgb{0.00,0.00,0.00}
      \put(1915,632){\rotatebox{-45}{\makebox(0,0)[l]{\strut{}$40$}}}%
      \colorrgb{0.00,0.00,0.00}
      \put(2309,632){\rotatebox{-45}{\makebox(0,0)[l]{\strut{}$60$}}}%
      \colorrgb{0.00,0.00,0.00}
      \put(2704,632){\rotatebox{-45}{\makebox(0,0)[l]{\strut{}$80$}}}%
      \colorrgb{0.00,0.00,0.00}
      \put(3099,632){\rotatebox{-45}{\makebox(0,0)[l]{\strut{}$100$}}}%
      \colorrgb{0.00,0.00,0.00}
      \put(3494,632){\rotatebox{-45}{\makebox(0,0)[l]{\strut{}$120$}}}%
      \colorrgb{0.00,0.00,0.00}
      \put(3888,632){\rotatebox{-45}{\makebox(0,0)[l]{\strut{}$140$}}}%
      \colorrgb{0.00,0.00,0.00}
      \put(4283,632){\rotatebox{-45}{\makebox(0,0)[l]{\strut{}$160$}}}%
    }%
    \gplgaddtomacro\gplfronttext{%
      \csname LTb\endcsname
      \put(198,1933){\rotatebox{-270}{\makebox(0,0){\strut{}Count}}}%
      \put(2704,154){\makebox(0,0){\strut{}Tokens per Played Wild Card}}%
      \csname LTb\endcsname
      \put(4125,2775){\makebox(0,0)[r]{\strut{}total=784708}}%
      \put(4125,2606){\makebox(0,0)[r]{\strut{} mean=23.96}}%
      \put(4125,2438){\makebox(0,0)[r]{\strut{} std=13.32}}%
      \put(4125,2270){\makebox(0,0)[r]{\strut{}bin width=7}}%
    }%
    \gplbacktext
    \put(0,0){\includegraphics{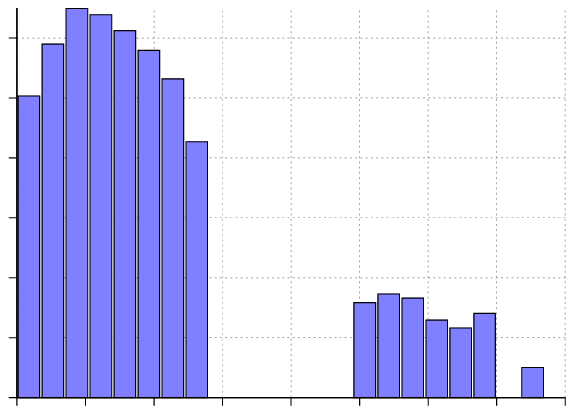}}%
    \gplfronttext
  \end{picture}%
\endgroup
}

\begin{table*}[tb]
  \centering
  \def\nsig{\makebox[0pt][l]{\textsuperscript{$\dagger$}}}
\begin{tabular}{l|cc|cc|cc|cc|cc|cc}
  & \multicolumn{2}{c|}{Likability} & \multicolumn{2}{c|}{Fluency} & \multicolumn{2}{c|}{Coherence} & \multicolumn{2}{c|}{\rougel} & \multicolumn{2}{c|}{\rougew} & \multicolumn{2}{c}{\storymetric} \\
  & top-$k$ & nuc & top-$k$ & nuc & top-$k$ & nuc & top-$k$ & nuc & top-$k$ & nuc & top-$k$ & nuc \\
  \hline

  Relevance & 0.51
            & 0.53
            & 0.28
            & 0.40
            & 0.55
            & 0.57
            & 0.52
            & 0.38
            & 0.50
            & 0.36
            & 0.51
            & 0.39 \\

  Likability & \multicolumn{2}{c|}{---}
             & 0.28
             & 0.38
             & 0.35
             & 0.55
             & 0.29
             & 0.34
             & 0.28
             & 0.31
             & 0.34
             & 0.35 \\

  Fluency & \multicolumn{2}{c|}{---}
          & \multicolumn{2}{c|}{---}
          & 0.54
          & 0.61
          & 0.11\nsig
          & 0.23
          & 0.10\nsig
          & 0.22
          & 0.13\nsig
          & 0.23 \\

  Coherence & \multicolumn{2}{c|}{---}
            & \multicolumn{2}{c|}{---}
            & \multicolumn{2}{c|}{---}
            & 0.27
            & 0.38
            & 0.24
            & 0.34
            & 0.25
            & 0.36 \\

  \rougel\ & \multicolumn{2}{c|}{---}
           & \multicolumn{2}{c|}{---}
           & \multicolumn{2}{c|}{---}
           & \multicolumn{2}{c|}{---}
           & 0.98
           & 0.98
           & 0.95
           & 0.93 \\

  \rougew\ & \multicolumn{2}{c|}{---}
           & \multicolumn{2}{c|}{---}
           & \multicolumn{2}{c|}{---}
           & \multicolumn{2}{c|}{---}
           & \multicolumn{2}{c|}{---}
           & 0.97
           & 0.94 \\
  \hline
\end{tabular}

  \caption{\storymetric\ correlates well with both \rougel\ and \rougew\ when
  removing stopwords.}
  \label{table:rouge-correlations}
\end{table*}

\section{Web Service}
\label{appendix:web-service}
Our web service is modular and allows easily adding new models. It consists of
a frontend service, which acts as a mediator between \storium\ and each backend
service responsible for serving model outputs. The frontend stores data in a
PostgreSQL database and provides a dashboard for viewing realtime ratings and
evaluation metrics. It also displays user comments, scene entry diffs based on
user edits, and Pearson's $r$ correlations among metrics and user ratings ---
all sortable per model. A new model can be served by simply implementing four
methods (\texttt{startup}, \texttt{shutdown}, \texttt{preprocess}, and
\texttt{generate}). The backend automatically installs all Python requirements
for serving a model and is agnostic to the underlying tensor library used.
Additionally, we follow the latest best practices, including the use of Docker
containers and the Asynchronous Server Gateway Interface
(ASGI)\footnote{FastAPI (\url{https://fastapi.tiangolo.com})},the latest Python
web standard, which allows for asynchronous programming using
\texttt{asyncio}.\footnote{\href{https://docs.python.org/3/library/asyncio.html}{\texttt{https://docs.python.org/3/library/}}
\href{https://docs.python.org/3/library/asyncio.html}{\texttt{asyncio.html}}}
We host the web service using an on-premise server with four 2080Ti GPUs.

\section{\storymetricfull}
\label{appendix:metric}
Recently, the discriminative power of \bleu\ has been called into question when
evaluating state-of-the-art machine translation systems, leading researchers to
investigate alternative evaluation metrics
\cite{Freitag2020BLEUMB,Sellam2020BLEURTLR}. Similarly, we question the use of
\rouge\ metrics for automatic evaluation of open-ended story generation. Using
our evaluation platform, we show that \storymetric\, improves upon \rouge\ in
the story generation domain. 

When evaluating story continuations, we cannot compare against an \textit{a
priori} gold standard. Rather, we consider the final published story a user
generates to be the gold standard, and thus evaluate models by how much text
the user retains. Using \rougel\ \textit{precision}, which simply computes
the ratio of the longest common subsequence (LCS) with the number of tokens in
the generated text, we can measure this quantity.

As highlighted by \citet{Lin2004ROUGE}, \rougel\ contains a subtle mismatch
with expectations, as the LCS does not consider \textit{locality} of matches
--- assigning equal weight to subsequences of the same length even when the
distance between matched words differs. Given a reference sequence $X$, the
following two candidate sequences $Y_1$ and $Y_2$ produce the same \rougel\
score (an underscore indicates a subsequence match):

\begin{equation*}
  \begin{split}
    X: [\mathrm{\underline{A}~\underline{B}~\underline{C}~\underline{D}~E~F~G}] \\
    Y_1: [\mathrm{\underline{A}~\underline{B}~\underline{C}~\underline{D}~H~I~K}] \\
    Y_2: [\mathrm{\underline{A}~H~\underline{B}~K~\underline{C}~I~\underline{D}}]
  \end{split}
\end{equation*}

\rougew\ tries to address this shortcoming by introducing a weighting which
favors subsequences with less separation. Sadly, for long texts, both \rougel\
and \rougew\ often favors long \textit{subsequences} of stopwords over
\textit{contiguous substrings}, a sign that a user clearly used part of the
output unchanged. While acceptable for short summaries, this is much less
appropriate for long-form open-ended text generation. Removing stopwords helps
alleviate the mismatch, so we do so in our comparison to \rouge\
(\tableref{rouge-comparison}), though the fundamental issue still remains. This
mismatch calls into question the ability of \rougel\ and \rougew\ to
distinguish among models with strong story generation capability.

\begin{table}[H]
  \centering
  \begin{tabular}{lcccc}
  & \multicolumn{2}{c}{Top-$k$} & \multicolumn{2}{c}{Nucleus} \\
   & Score & Count & Score & Count\\
  \hline
  \multicolumn{1}{l|}{\rougel}       & \textbf{28.61} & 174 & 20.66 & 178\\
  \multicolumn{1}{l|}{\rougew}       & \textbf{20.73} & 174 & 13.80 & 178\\
  \multicolumn{1}{l|}{\storymetric}  & \textbf{15.63} & 174 & 9.86 & 178\\
  \hline
\end{tabular}

  \caption{\storymetric\ produces lower scores on average than \rougel\ or
  \rougew.}
  \label{table:rouge-comparison}
\end{table}

Our new metric, \storymetricfull\ (\storymetric), is based on a diff-like
approach. We begin by applying the same text preprocessing as \rouge.
Afterwhich, we find the longest contiguous substring, then use it as a pivot to
divide the remaining string into two halves (excluding the pivot), and
recursively repeat the process in each half.\footnote{We use
\texttt{SequenceMatcher} from Python's \texttt{difflib}:
\href{https://docs.python.org/3/library/difflib.html}{\texttt{https://docs.python.org/3/library/}}
\href{https://docs.python.org/3/library/difflib.html}{\texttt{difflib.html}}}
We then only consider substrings with at least one non-stopword as
\highlightinline{matches} (careful scrutiny of \figureref{example_cards}
reveals an unmatched stopword \textit{it}). Subsequently, we compute
precision, recall, and F1 identically to \rouge.

\tableref{rouge-correlations} shows \storymetric\ correlates with user
judgments approximately similarly to \rouge\ metrics, while correlating
strongly with both metrics. Additionally, \storymetric\ produces lower scores
on average compared to \rouge\ (\tableref{rouge-comparison}). Taken in
combination, these insights indicate \storymetric\ is better capable of
discerning differences among the strong story generation models of the
future, as it provides more stark evaluations while still correlating well with
human judgments.

\end{document}